\documentclass[pdflatex,sn-basic, Numbered]{sn-jnl}

\usepackage{graphicx}%
\usepackage{multirow}%
\usepackage{amsmath,amssymb,amsfonts}%
\usepackage{amsthm}%
\usepackage{mathrsfs}%
\usepackage[title]{appendix}%
\usepackage{xcolor}%
\usepackage{textcomp}%
\usepackage{manyfoot}%
\usepackage{booktabs}%
\usepackage{algorithm}%
\usepackage{algorithmicx}%
\usepackage{algpseudocode}%
\usepackage{listings}%
\usepackage[utf8]{inputenc}
\usepackage[T1]{fontenc}
\usepackage{subcaption}
\usepackage{graphicx}%
\usepackage{multirow}%
\usepackage{amsmath,amssymb,amsfonts}%
\usepackage{amsthm}%
\usepackage{mathrsfs}%
\usepackage[title]{appendix}%
\usepackage{xcolor}%
\usepackage{textcomp}%
\usepackage{manyfoot}%
\usepackage{booktabs}%
\usepackage{algorithm}%
\usepackage{algorithmicx}%
\usepackage{algpseudocode}%
\usepackage{listings}%
\usepackage{comment}
\usepackage{bm}
\usepackage{todonotes}
\usepackage{natbib}
\usepackage{multirow}
\usepackage{tabularx}
\usepackage{longtable}
\usepackage{tcolorbox}
\tcbuselibrary{skins, breakable}
\usepackage{pgf}
\usepackage{tikz}

\theoremstyle{thmstyleone}%
\theoremstyle{thmstyletwo}%

\theoremstyle{thmstylethree}%

\begin{document}

\title[Assessing VLM Reliability for Medical Image Quality Evaluation Under Corruption and Bias]{Assessing VLM Reliability for Medical Image Quality Evaluation Under Corruption and Bias}


\author[1,2,*]{Sofiane Ouaari}
\author[1,2]{Kevin Vorwalder}
\author[1,2]{Nico Pfeifer}
\affil[1]{Methods in Medical Informatics, Department of Computer Science, University of Tuebingen, Germany}
\affil[2]{Institute for Bioinformatics and Medical Informatics (IBMI), University of Tuebingen, Germany}

\affil[*]{sofiane.ouaari@uni-tuebingen.de}


\abstract{\textbf{Background}: Vision-Language Models (VLMs) are increasingly applied in medical tasks such as pathology description, report generation, and visual question answering. Medical Image Quality Assessment (MIQA) supports diagnostic accuracy and patient safety by determining whether images meet the standards required for clinical decision-making. Automating MIQA with VLMs may reduce workload, but their behavior under real-world conditions, where images may be degraded or textual context may affect judgments, should be further explored before deployment.

\textbf{Methods}: We benchmark VLMs on medical image quality using the MediMeta-C dataset zero-shot across seven corruption types and five severity levels. We evaluate sensitivity to degradation patterns, the effect of corruptions on embedding geometry, and whether textual attributes (demographics, expertise, infrastructure, institution) alter scores.

\textbf{Results}: Across 16 VLMs and seven modalities, pixelation produced the largest score reductions (mean -20.58\%, up to -34.4\% for OCT), whereas brightness had limited effect (-0.81\%). Embedding displacement was associated with score changes. Same-family models showed correlations of 0.67-0.83; some produced increases up to +31\% for corrupted mammography. Textual attributes affected scores: institutional prestige raised them +17.15\%, and equipment age lowered them -14.7\%. The largest changes were +95.62\% (\texttt{InternVL-8B}) and -37.7\% (\texttt{MedGemma}).

\textbf{Conclusions}: Current VLMs show limitations for medical image quality assessment. Pixelation, a privacy-preserving transformation, reduces performance, indicating a trade-off between patient privacy and reliability. Sensitivity to contextual metadata indicates limited objectivity and marks metadata as a privacy and bias source. Privacy protection and objective quality assessment are related requirements for use.

}

\keywords{Visual Language Models, Medical Imaging, Medical Image Quality Assessment, Image Artifacts, Biased Evaluation}



\maketitle

\section*{Introduction}\label{sec1}
Visual Language Models (VLMs) represent a significant evolution of Large Language Models (LLMs), enabling multimodal reasoning by seamlessly integrating visual and textual information. These models have demonstrated broad applicability across diverse domains, from general-purpose vision-language understanding to specialized applications in robotics and document analysis \citep{ghosh2024exploring, shinde2025survey, awais2025foundation}. The medical imaging community has particularly embraced VLMs, leveraging their capabilities for critical tasks including automated radiology report generation from chest X-rays \citep{tanno2025collaboration}, medical visual question answering \citep{hartsock2024vision}, disease classification and image retrieval \citep{wang2022medclip, li2025medbridge}, and anatomical slice selection \citep{wang2024enhancing}.

Healthcare systems increasingly adopt AI-driven automation for diagnostic and administrative workflows, VLMs have emerged as key components in this transformation. However, a fundamental challenge remains partially unaddressed: the reliability and trustworthiness of these models when deployed in safety-critical medical contexts \citep{belisle2024we, yim2024err}. Among the various factors affecting model performance, image quality assessment stands out as particularly crucial as corrupted or degraded medical images can lead to diagnostic errors with severe consequences for patient care.

Recent studies have further analyzed and evaluated VLM capabilities for medical imaging applications. MedQ-Bench \citep{liu2025medq} introduced a valuable perception-reasoning paradigm for assessing whether VLMs can identify and reason about quality degradations through structured question-answering, demonstrating that models possess preliminary perceptual abilities while highlighting areas for improvement. This approach provides rich insights into VLM understanding of specific artifact types through language-based evaluation. Similarly, \citealt{cheng2025understanding} investigated VLM robustness to artifacts during disease detection tasks, revealing how performance varies across different corruption types and establishing important benchmarks for diagnostic robustness. Their work advances the understanding relative to how VLMs handle degraded images in classification scenarios. Building on these prior work, important questions remain regarding VLM reliability for continuous quality assessment and how textual contextual information might influence these judgments. Furthermore, the intersection of privacy-preserving image transformations and VLM performance, as well as the potential for metadata to introduce systematic biases, requires a deeper investigation for safe clinical deployment.

In this study, we present a comprehensive benchmark evaluating state-of-the-art VLMs on medical image quality assessment using the MediMeta-C \citep{imam2025robustness} dataset, which contains both clean medical images and systematically corrupted versions spanning seven corruption types at five severity levels. Our investigation addresses three key research questions: First, we analyze how VLMs score image quality across different corruption types and severities, revealing their sensitivity to various degradation patterns. Second, we examine the embedding space geometry through PCA visualization and embedding distance analysis to understand how corruption-induced distribution shifts affect model representations. Finally, we investigate the influence of textual bias in prompts; specifically, whether incorporating metadata affects quality assessments, uncovering vulnerabilities in prompt-based evaluation paradigms. Through this systematic analysis, we aim to establish baselines for medical image quality assessment with VLMs and identify critical failure modes that must be addressed prior to clinical deployment.

\begin{figure}[h]
\centering
 \begin{subfigure}[b]{0.72\textwidth}
         \centering
         \includegraphics[width=\textwidth]{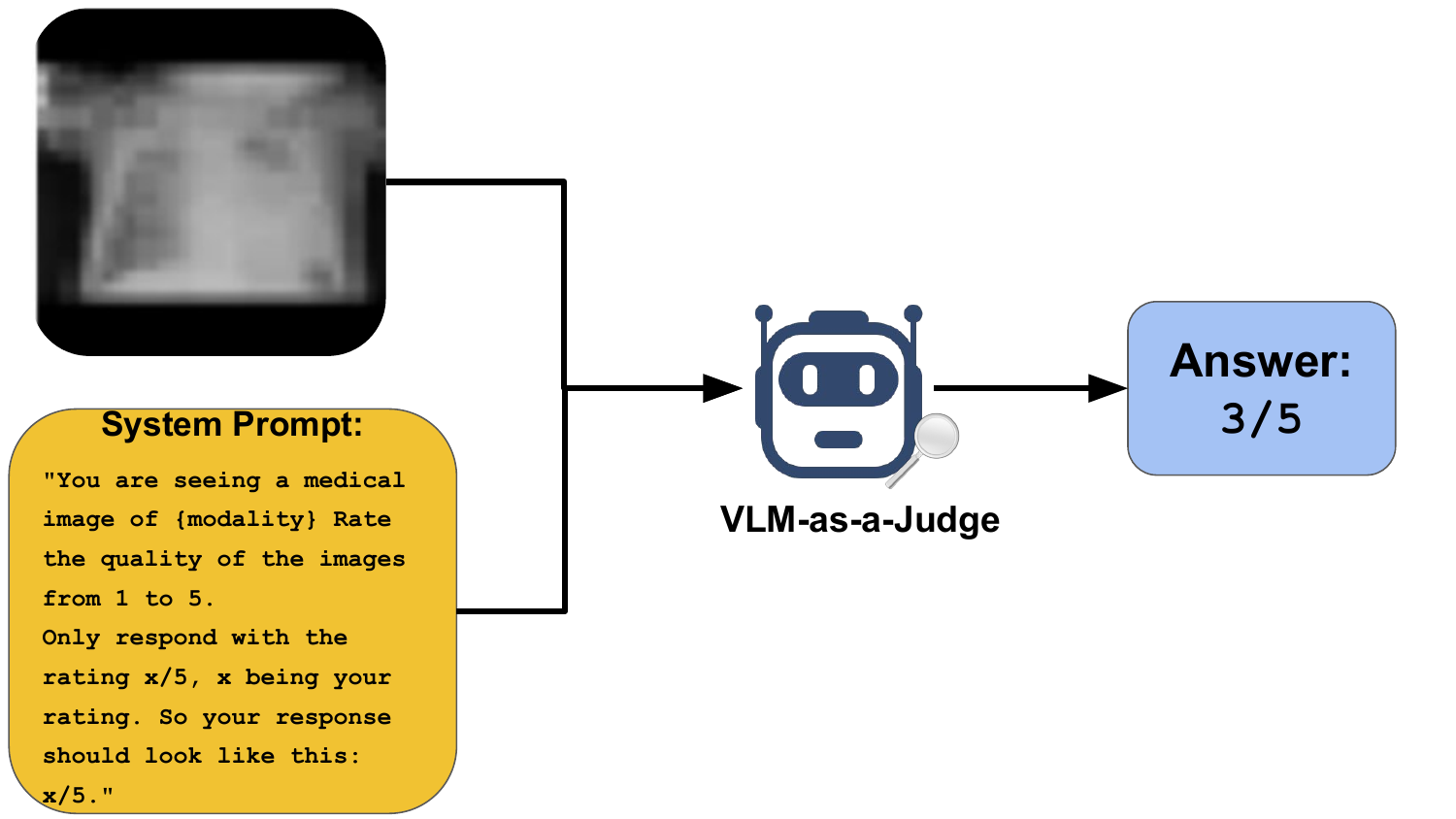}
         \caption{Pipeline of analyzing the behavior of VLM-as-a-Judge for medical image quality assessment on corrupted images of MediMeta-C}
         \label{fig:vlm_as_judge_corruption}
     \end{subfigure}
  \hfill
     \begin{subfigure}[b]{0.72\textwidth}
         \centering
         \includegraphics[width=\textwidth]{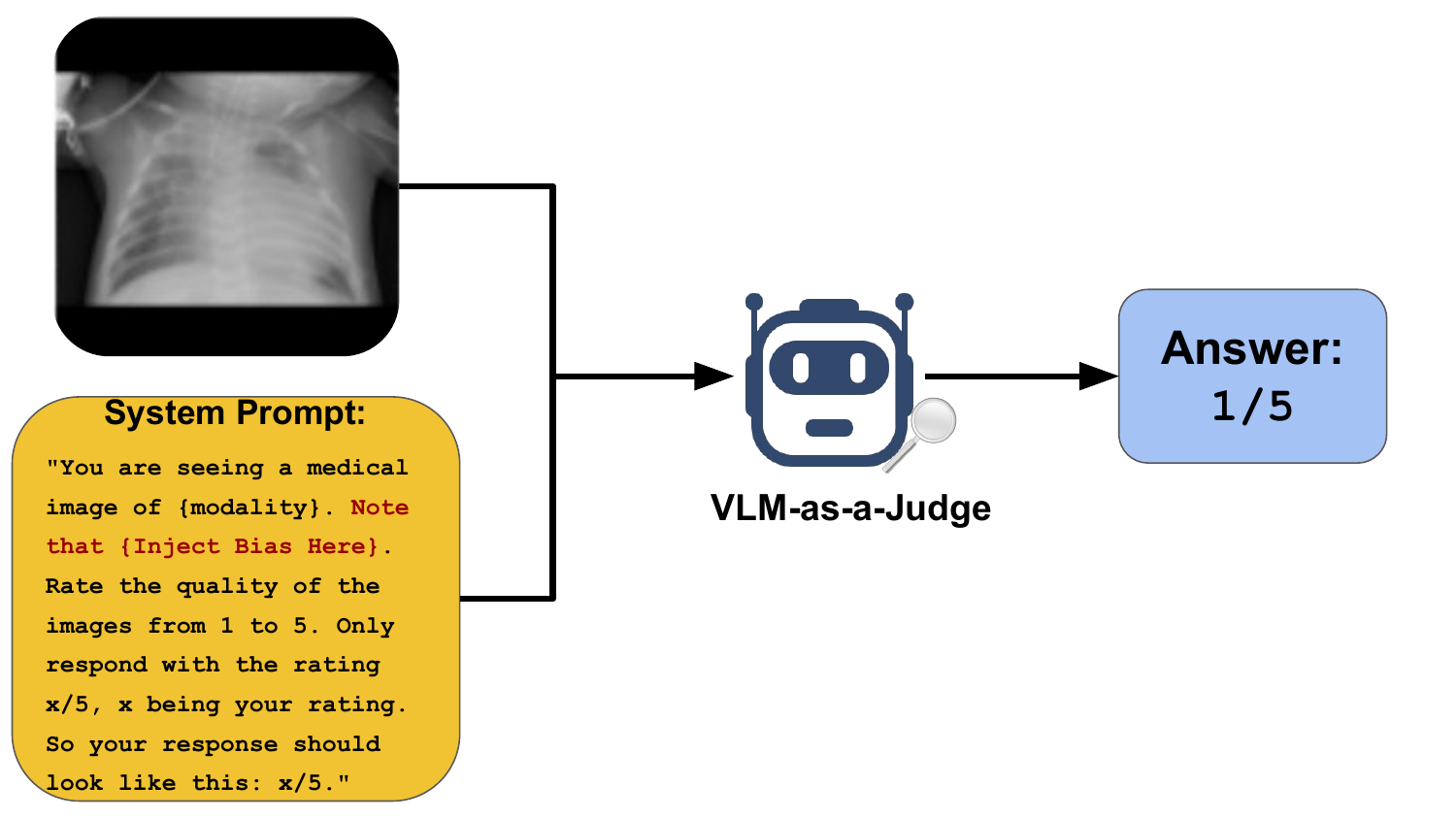}
         \caption{Pipeline of analyzing the behavior of VLM-as-a-Judge on clean images with added text bias}
         \label{fig:vlm_as_a_judge_bias}
     \end{subfigure}

\caption{Experimental pipelines for evaluating VLM-as-a-Judge behavior in medical image quality assessment.}
\label{fig:overall_architecture}
\end{figure}

\section*{Background}\label{sec2}

\subsection*{Visual Language Models (VLM)}

A VLM is a multimodal neural network based on transformers \citep{vaswani2017attention} that processes paired text-image inputs to generate textual outputs.
Given a dataset $\mathcal{D} = \{(t, i)\}$ where $t \in \mathcal{T}$ represents text and $i \in \mathcal{I}$ represents images. 
A VLM consists of mainly a text encoder $E_t$ and a vision encoder $E_v$. 
The text encoder $E_t$ maps text $t$ to a textual feature representation of dimension $d_t$: $E_t : \mathcal{T} \rightarrow \mathbb{R}^{d_t}$.
The vision encoder on the other hand maps an image $i$ to a visual feature representation of dimension $d_v$: $E_v : \mathcal{I} \rightarrow \mathbb{R}^{d_v}$. 

In order to align $E_t$ and $E_v$, an embedding projector $P$ is used on the vision embedding to ensure dimensional compatibility with the text embedding ${P}(E_v(i)) \in \mathbb{R}^{d_t}$. 
The VLM then produces $t_{\text{out}}$, a textual response conditioned on both visual and textual inputs.

\subsection*{VLM-as-a-Judge}
VLM-as-a-judge is an application pattern and refers to the use of VLMs as automated evaluators to assess the quality, accuracy, and relevance of outputs, particularly in tasks involving visual and textual modalities. This paradigm extends the established "LLM-as-a-Judge" \citep{gu2024survey} framework into multimodal domains, enabling systematic evaluation of vision-language tasks without extensive human annotation.
Formally let us consider a VLM model $\mathcal{M}$ which takes as an input a text query $t$ and an image $i$ and outputs an evaluation score $s = \mathcal{M}(t,i)$, where $s$ can be a discrete scoring $s \in \mathbb{Z}$ or a continuous scoring $s \in \mathbb{R}$.
\subsection*{Medial Image Quality Assessment}
Medical imaging as a modality represents one of the most abundant sources of clinical data generated and archived by healthcare institutions. Despite their widespread use across diverse storage formats, these images are frequently compromised by artifacts and noise that degrade image quality, thereby potentially undermining diagnostic accuracy and prognostic reliability. The Medical Image Quality Assessment (MIQA) \citep{chow2016review, ma2024rad} addresses this critical challenge by systematically evaluating image integrity to ensure clinical utility for diagnostic and therapeutic decision-making. MIQA encompasses methodologies that quantify key parameters including contrast, spatial resolution, signal-to-noise ratio, artifact prevalence, and geometric distortion. Robust quality assessment is critical to maintaining diagnostic accuracy and improving patient care.

\section*{Methods}\label{sec3}

We study the reliability of using non-medical and medical VLMs in a zero-shot fashion as a way to automatically score and evaluate the overall quality of a medical image (\autoref{fig:vlm_as_judge_corruption}). 
To address this we benchmarked and evaluated a total of 16 open-source VLMs available on \textit{Hugging Face}\footnote{https://huggingface.co/}, 11 of which are general purpose and 5 being medically-oriented. We run our experiments on the MediMeta-C \citep{imam2025robustness} a dataset which includes 5 severity levels on 7 modalities.

\subsection*{Data}
For our study, we use MediMeta-C \citep{imam2025robustness}, a corruption benchmark derived from the MediMeta \citep{woerner2025comprehensive} dataset. MediMeta-C systematically applies various perturbations across multiple medical imaging datasets to evaluate model robustness under realistic image degradation conditions.
The dataset includes the following main modalities: Cell Microscopy, Breast Imaging, Chest X-ray, Funduscopy, and Retinal OCT. Cell Microscopy includes two subtypes: Acute Myeloid Leukemia (AML) and Peripheral Blood Cells (PBC). Similarly, Breast Imaging comprises Mammography Calcification (MAMMO\_CALC) and Mammography Masses (MAMMO\_MASS).
MediMeta-C incorporates seven corruption types that simulate common imaging artifacts: Motion Blur, Zoom Blur, Gaussian Noise, Impulse Noise, Brightness, Contrast, and Pixelation. These corruptions represent realistic image quality degradations encountered in clinical settings, enabling comprehensive evaluation of model performance under diverse conditions. A total of five corruption levels were used in order to vary the amount of perturbation added to the image from low to high. 

\subsection*{Evaluation Pipeline}
\label{sec:evaluation_pipeline}

To establish a robust baseline for comparison and understand how VLMs assess medical image quality without perturbations, we first examine their behavior in a zero-shot setting on clean images. This baseline, used as a reference, enables us to later on properly quantify the impact of different corruption types and severity levels on perceived image quality.

We begin by evaluating how VLMs rate the image quality of clean images $\mathcal{I}^{\text{clean}}$ from the original MediMeta dataset \citep{woerner2025comprehensive}. Formally, we obtain quality scores as
\begin{equation}
    \mathcal{S}^{clean} = \{\mathcal{M}(t, i) \mid i \in \mathcal{I}^{clean}\},
\end{equation}
where $t$ denotes the evaluation prompt, $i \in \mathcal{I}^{\text{clean}}$ represents an image from the set of clean, unperturbed images, and $\mathcal{M}$ is the VLM performing the assessment. The resulting score set is defined as $\{s \mid s \in \{1, 2, 3, 4, 5\}\}$, where the discrete rating scale $s \in \{1, 2, 3, 4, 5\}$ matches the perturbation severity levels employed in MediMeta-C, ensuring consistent comparison across conditions.

Let $\mathcal{I}_{k}^{c}$ denote the image set for corruption type $c \in \mathcal{C}$ at severity level $k \in \{1, 2, 3, 4, 5\}$, where $\mathcal{C}$ represents the set of all corruption types. We perform the quality assessment on each image set, yielding the corresponding score set
\begin{equation}
    \mathcal{S}_{k}^{c} = \{\mathcal{M}(t, i) \mid i \in \mathcal{I}_{k}^{c}\},
\end{equation}
which aggregates all VLM quality ratings for images corrupted with type $c$ at level $k$.

In order to observe which corruption type $c$ effects the judgment of the VLMs, we compute the percentage change as follows: 
\begin{equation}
    \mathcal{P}^{c} = \frac{\bar{s}^c - \bar{s}^{clean}}{\bar{s}^{clean}}\
\end{equation}

with $\bar{s}^{clean}$ and $\bar{s}^c$ being the average scores of $\mathcal{S}^{clean}$ and $\mathcal{S}^{c}$ respectively. 

\subsection*{Used VLMs}
In this study, we focus exclusively on open-source and freely available VLMs to promote reproducibility and facilitate adoption by the broader research community. Our model selection strategy prioritizes diversity across three dimensions: model family, parameter scale, and domain specialization. This yields a comprehensive set of 16 VLMs, comprising 11 general-purpose models and 5 medical-specific variants.

Diversifying across model families and parameter scales is essential for establishing robust and generalizable benchmarking results. First, different model families employ distinct architectural designs, training objectives, and data handling strategies, which can lead to fundamentally different learned representations. By evaluating multiple families, we avoid drawing conclusions that may be artifacts of a single architectural paradigm and instead identify corruption sensitivities that reflect broader limitations of current VLM approaches. Second, examining models across various parameter scales from compact models suitable for resource-constrained deployment to large-scale models representing state-of-the-art capabilities enables us to investigate whether robustness properties scale predictably with model capacity or whether specific vulnerabilities persist regardless of scale. This is particularly important for understanding whether robustness can be achieved simply through scaling or requires targeted architectural and training innovations. Third, evaluating both general and domain-specialized models allows us to assess whether domain adaptation improves a model's robustness against specific corruptions, or if a narrower training distribution inadvertently creates new weaknesses.

For this study the following VLMs are being evaluated: Medgemma 4B/27B \citep{sellergren2025medgemma}, Lingshu Medical 7B/32B \citep{xu2025lingshu}, Qwen2.5-VL 3B/7B/32B \citep{bai2025qwen2}, Qwen3-VL 2B/4B/8B \citep{yang2025qwen3}, LLava-v1.6-mistral-7B, Llama3-llava-next-8B, LLava-med-v1.5-mistral-7B \citep{liu2023visual, li2023llava, liu2024improved} and InternVL 3.5-2B/8B/14B \citep{wang2025internvl3}.

\subsection*{Embedding Analysis}

\begin{figure*}
    \includegraphics[width=\linewidth]{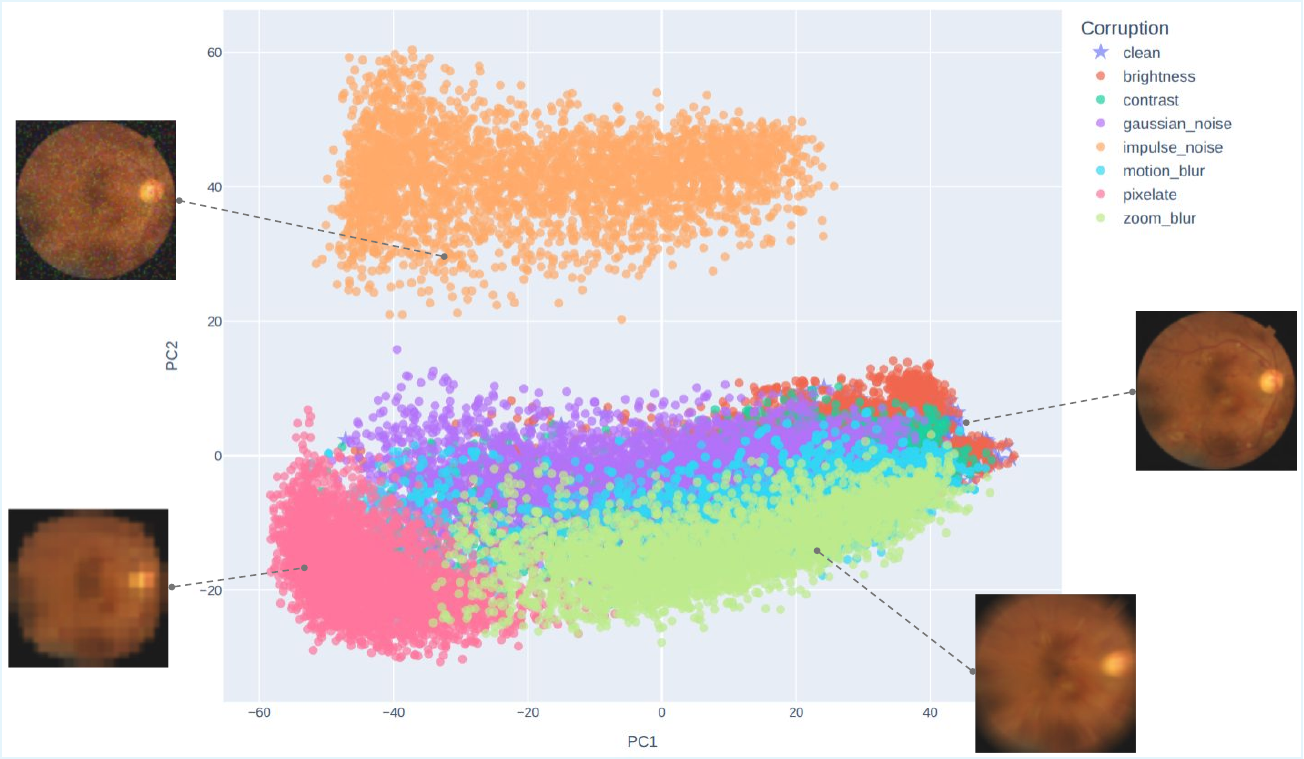}
    \caption{Two-dimensional PCA visualization of embeddings from Llava-v1.6-mistral-7B for fundus images in the MediMeta-C dataset. Each point represents an image embedding, with clean images serving as the reference distribution. The plot illustrates the separation between clean and corrupted samples in the learned embedding space}
    \label{fig:embedding_vlm}
\end{figure*}

Analysis of hidden state embeddings has been demonstrated to be an effective method for understanding the behavior of vision-language models \citep{sun2024general,masry2025alignvlm,papadimitriou2025interpreting}. 

Let $\mathcal{E} = \{e\}$ denote the set of joint embedding representations corresponding to an image dataset $\mathcal{I}$. An example of 2D embeddings of Fundus images is shown in \autoref{fig:embedding_vlm}.

For each corruption type $c \in \mathcal{C}$, we quantify the geometric displacement in embedding space by computing the Euclidean distance between centroids:
\begin{equation}
    d^{c} = \left\|\bar{e}^{c} - \bar{e}^{\text{clean}}\right\|_{2}
    \label{eq:distance_embeddings}
\end{equation}
where the embedding centroids are defined as:
\begin{equation}
    \bar{e}^{\text{clean}} = \frac{1}{N}\sum_{i=1}^{N} e_i^{\text{clean}}, \quad \bar{e}^{c} = \frac{1}{N}\sum_{i=1}^{N} e_i^{c}
\end{equation}
with $\{e^{\text{clean}}\}$ and $\{e^{c}\}$ denoting the embeddings of clean image $\mathcal{I}^{\text{clean}}$ and its corrupted variant $\mathcal{I}^{c}$, respectively.

This metric enables us to investigate the relationship between geometric shifts in the latent space and model output change. Specifically, we examine whether corruptions that induce larger embedding displacements systematically correspond to greater evaluation variation.

\subsection*{Bias effect analysis}

In addition to analyzing visual robustness through the image corruption types and severity levels presented above, we extend our investigation to the textual modality of VLMs. Specifically, we examine how medical image quality assessment (MIQA) performance is influenced by the incorporation of metadata into the system prompt during evaluation. This dual-modality analysis is motivated by the fact that in clinical practice, medical images are rarely interpreted in isolation and are typically accompanied by contextual information such as patient or doctor demographics, hospital location, imaging protocols, and acquisition parameters. By systematically varying the metadata provided in the prompt as shown in \autoref{fig:vlm_as_a_judge_bias}, we aim to characterize whether and how such contextual information modulates VLM quality assessments, and whether this contextual sensitivity differs across corruption types or model architectures. Understanding this interaction between text and image is essential for deploying VLMs in real-world medical settings, where the availability and quality of accompanying metadata may vary substantially across institutions, imaging modalities, and clinical workflows.

To investigate how textual context influences VLM-based medical image quality assessment, we introduce controlled biases through metadata manipulation in the system prompt. Our approach builds upon the bias taxonomy proposed in CALM~\citep{ye2024justice}, originally developed for evaluating LLM-as-a-Judge frameworks, which we adapt and extend to the vision-language domain with medical-specific considerations.
\begin{figure}
\begin{tcolorbox}[
    breakable,
    enhanced,
    colback=green!5,
    colframe=green!50!black,
    title=Baseline Prompt,
    fonttitle=\bfseries,
    width=\textwidth,]
\begin{quote}
\ttfamily
You are seeing a medical image of ['modality']. Rate the quality of the images from 1 to 5. Only respond with the rating x/5, x being your rating. So your response should look like this: x/5.
\end{quote}
\end{tcolorbox}
\caption{Baseline prompt template and structure used to obtain the baseline score $\mathcal{S}^{\text{clean}}$ and to evaluate images across multiple corruption types and severity levels}
\label{quote:baseline_prompt}
\end{figure}
\begin{figure*}
    \centering
    \includegraphics[width=\linewidth]{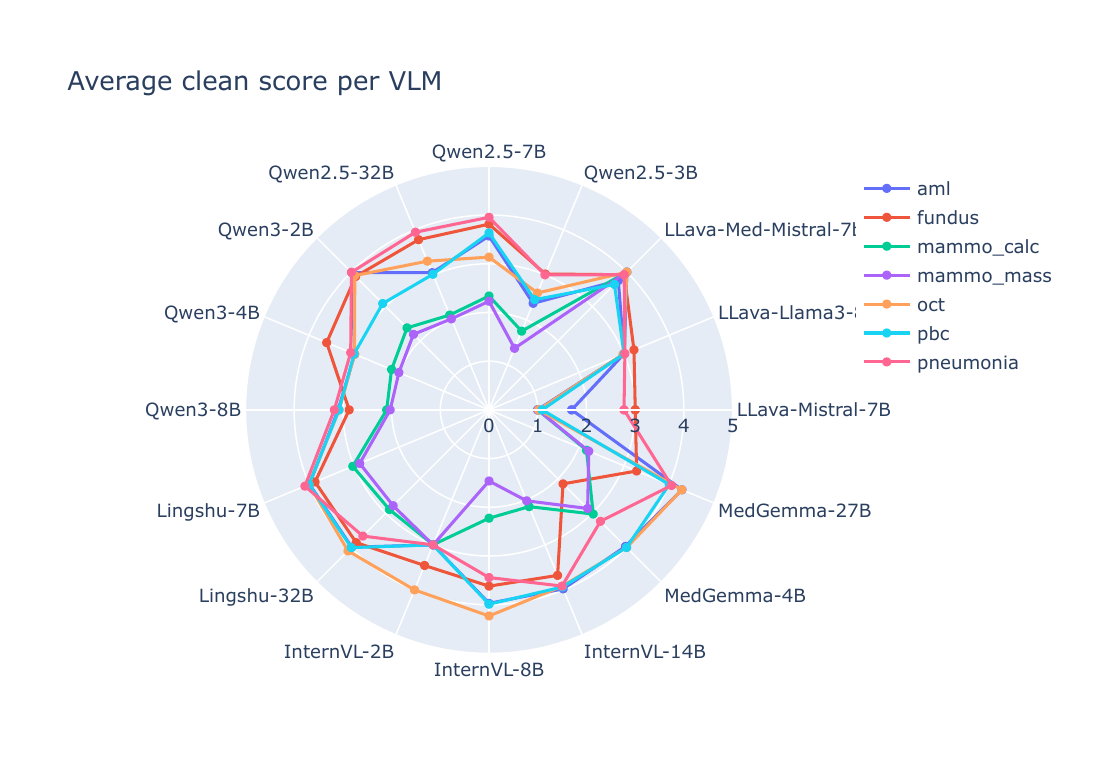}
    \caption{Comparative radar chart visualization of average image quality assessments across vision-language models on clean images. Scores are rated on a five-point scale, with results shown for each model type.}
    \label{fig:radar_chart_clean_images}
\end{figure*}

We define a textual bias as any non-visual contextual information that may systematically influence a model's quality assessment despite being irrelevant to the actual image quality. We consider four bias categories, each designed to evaluate whether VLMs exhibit sensitivity to contextual factors that should not affect technical image quality assessment:

\textbf{Expertise Bias:} such a bias tests whether models assign different quality scores based on the skill level of the image acquisition operator. We compare prompts stating the image was \textit{``taken by a world-renowned medical expert''} versus \textit{``taken by a medical student''}. An unbiased model should produce identical assessments regardless of operator expertise, as technical image quality is determined by equipment parameters and patient factors, not operator credentials.

\textbf{Infrastructure Bias:} Examines sensitivity to equipment prestige by comparing \textit{``taken from a high-tech device''} against \textit{``taken from an old device''}. While equipment age may correlate with image quality in practice, the bias manifests when models overweight this context beyond what the visual evidence supports taking into consideration that the same clean image is being evaluated.

\textbf{Institution Bias:} Investigates whether institutional reputation affects assessments through prompts specifying \textit{``this was taken in a high world class medical institution''} versus \textit{``taken in a local medical institution''}. This tests whether models exhibit systematic preferences for images purportedly from prestigious institutions, independent of actual visual quality.

\textbf{Demographic Bias:} Probes for discriminatory associations related to physician race by varying whether the image was \textit{``taken by a white doctor''} or \textit{``taken by a black doctor''}. Any quality assessment difference based solely on this demographic information would indicate problematic learned associations that could perpetuate healthcare disparities.

This analysis is critical for clinical deployment, as VLMs must provide objective assessments resistant to contextual effects. Bias sensitivity could lead to systematic over or under estimation of image quality based on metadata rather than visual content, potentially compromising diagnostic workflows.

 \section*{Results}\label{sec5}
We established a reference baseline $\mathcal{S}^{\text{clean}}$ by systematically evaluating VLM performance on uncorrupted medical images from the MediMeta dataset. This baseline serves as the foundation for subsequent comparative analyses, enabling precise quantification of how image quality assessments deviate under controlled perturbations across different corruption types, severity levels, and textual bias conditions.

\begin{figure}[h]
     \centering
     \begin{subfigure}[b]{0.85\textwidth}
         \centering
         \includegraphics[width=\textwidth]{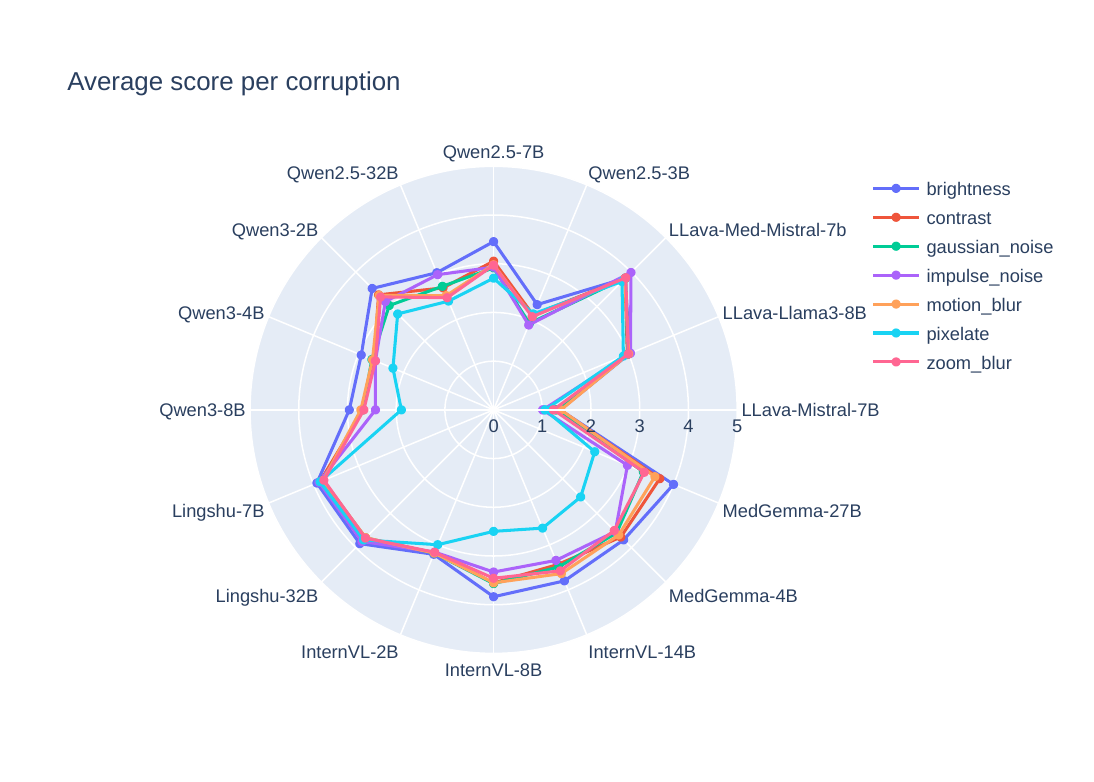}
         \caption{Comparative radar chart visualization of average image quality assessments across vision-language models depending on corruption type.}
         \label{fig:corruption_radar_chart}
     \end{subfigure}
  \hfill
     \begin{subfigure}[b]{0.85\textwidth}
         \centering
         \includegraphics[width=\textwidth]{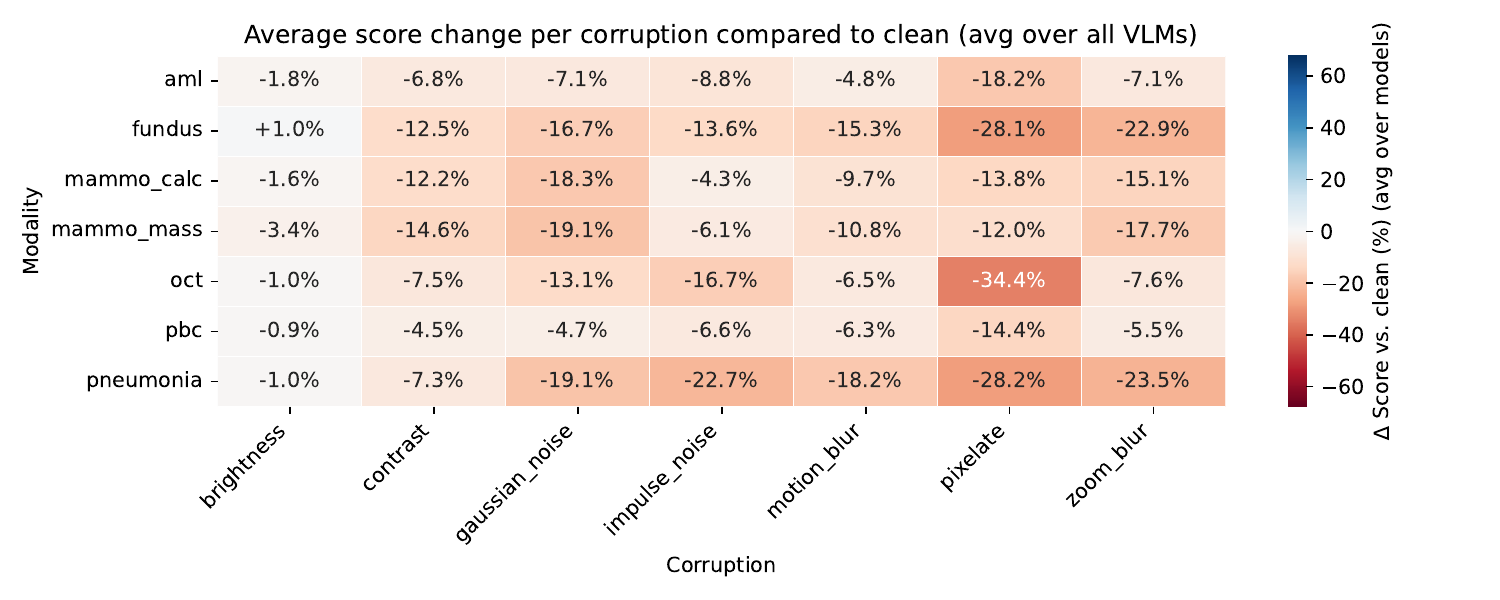}
         \caption{Comparative heatmap showing the average percentage change $\mathcal{P}^{c}$ per corruption type $c$ and medical image modality.}
         \label{fig:average_heatmap_corruption}
     \end{subfigure}

     \caption{Corruption types analysis per VLM and modality}
     \label{fig:corruption_types_analysis}
\end{figure}

The exact prompt being used is shown in \autoref{quote:baseline_prompt}. The average of quality scores assigned by each VLM to clean images is presented in \autoref{fig:radar_chart_clean_images}. Substantial difference in baseline scoring behavior was observed across models, with mean quality scores ranging from 1.523 to 3.994, on a scale between 1 to 5. Specifically, \texttt{LLava-Mistral-7B} exhibited the most conservative scoring profile, assigning the lowest average quality score of 1.523 across all imaging modalities. In contrast, \texttt{MedGemma-27B} demonstrated the highest average baseline score of 3.994, suggesting a more lenient quality assessment tendency or potentially superior alignment with uncorrupted medical image characteristics.

Distinct scoring patterns emerged across different medical imaging modalities. Mammography images both mass detection (MAMMO\_MASS) and calcification detection (MAMMO\_CALC) sequences received lower quality assessments, with mean scores of 2.415 and 2.548, respectively. This grading may reflect the inherently subtle contrast characteristics and fine-detail requirements of mammographic interpretation, or potentially indicate VLM limitations in processing mammographic features.
Conversely, optical coherence tomography (OCT) and chest radiography for pneumonia detection (PNEUMONIA) received substantially higher average scores of 3.644 and 3.544, respectively.
Notably, \texttt{MedGemma-27B} achieved the highest modality-specific average scores, reaching 4.286 for both OCT and acute myeloid leukemia (AML) microscopy images. This performance suggests that domain-specialized medical VLMs may exhibit enhanced sensitivity to the characteristic features of certain imaging modalities, particularly those with well-defined pathological signatures.

Aggregated analysis comparing medical versus non-medical VLMs (Supplementary Material, Figure 1a) revealed that Medical VLMs assigned elevated scores to clean images compared to their general-purpose counterparts, suggesting that domain-specific pretraining enhances the models' ability to recognize the quality characteristics of medical imaging data.

An examination of color effects (Supplementary Material, Figures 1b and 1c) demonstrated that colored images received marginally higher quality scores compared to grayscale images across all evaluated VLMs.

\subsection*{Image Corruption Scoring Change Analysis}

Following the baseline evaluation on clean images, we analyzed the quality scores across different corruption types in MediMeta-C at five severity levels. Each severity level controls the magnitude of degradation applied to the images. ~\autoref{fig:corruption_radar_chart} reveals that \textit{pixelate} corruption produced the largest reduction in quality scores across all VLMs and modalities, with an average score of 2.647, representing a decrease of -20.58\% relative to clean image scores. In contrast, \textit{brightness} had the smallest impact, yielding an average score of 3.306, corresponding to a minor reduction of -0.81\% compared to baseline. This  indicate that VLMs are more sensitive to corruptions that obscure fine details and reduce image sharpness, such as pixelation, than to changes in overall image intensity that preserve structural information, such as brightness variations.

To further examine this sensitivity, we calculated the percentage change in scores for each corruption type $c$ per image modality. The heatmap in ~\autoref{fig:average_heatmap_corruption} presents the average percentage change $\mathcal{P}^{c}$ per modality and across all VLMs. \textit{Pixelate} applied to OCT images resulted in the most pronounced score reduction, averaging -34.4\% across all evaluated VLMs. This modality-specific reduction likely reflects the importance of layer boundaries and fine structural details in OCT interpretation, which pixelation severely degrades.
Examining individual VLM responses across modalities and corruption types revealed notable patterns in model behavior. \texttt{MedGemma-27B} demonstrated the most consistent sensitivity to image degradation, showing the largest score reductions when comparing corrupted to clean images. The model's most substantial response was a -67.7\% decrease for pixelated FUNDUS images (Supplementary Material Figure 18), suggesting strong alignment between its quality assessment criteria and clinically relevant image features in retinal imaging.
However, we observed unexpected scoring patterns for certain model-corruption-modality combinations. Some VLMs assigned higher quality scores to corrupted images than to their clean counterparts. For instance, \texttt{InternVL-14B} increased scores by +31.4\% when \textit{impulse noise} was applied to MAMMO\_CALC images (Supplementary Material Figure 14). Similarly, \texttt{InternVL-8B} showed a +30.8\% score increase for the same corruption type on MAMMO\_MASS images (Supplementary Material Figure 13). These counterintuitive responses suggest that certain noise patterns may be misinterpreted as clinically relevant features, particularly in mammography where calcifications and masses can have textural properties that superficially resemble certain corruption artifacts. 
\begin{figure}[h]
    \centering
    \includegraphics[width=\linewidth]{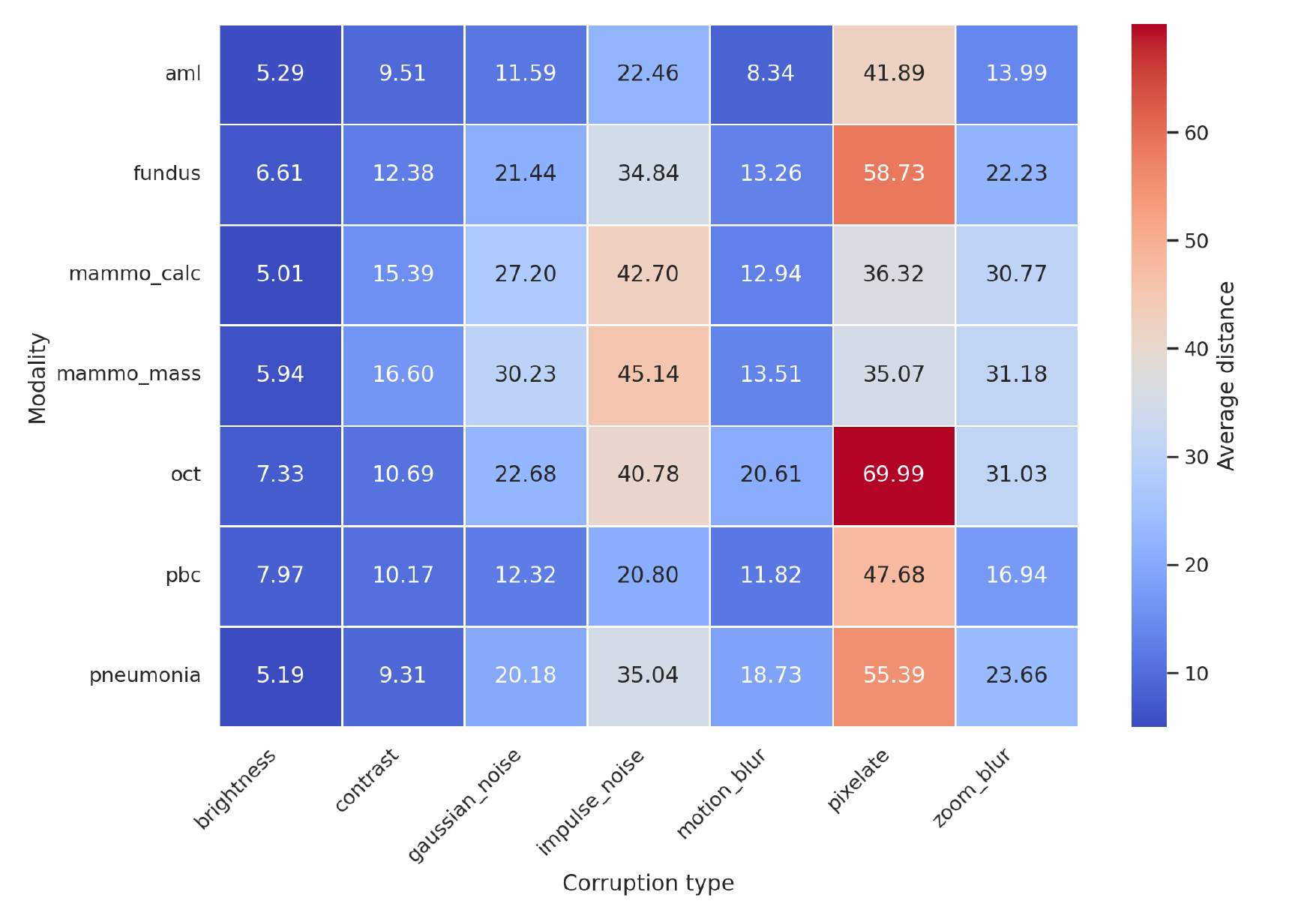}
    \caption{Average distance $d^{c}$ per corruption type $c$ for every modality}
    \label{fig:embedding_distance_heatmap}
\end{figure}

To better understand VLM behavior across different corruption types and explain the observed variations in quality score changes, we computed the distance $d^{c}$ between the centroid of clean image embeddings $\bar{e}^{\text{clean}}$ and the centroid of embeddings $\bar{e}^{c}$ for each corruption type
$c$, as defined in ~\autoref{eq:distance_embeddings}. This analysis aims to quantify how each corruption type shifts the representation space learned by the VLMs and to test whether these shifts correlate with the observed changes in quality assessment.
~\autoref{fig:embedding_distance_heatmap} reveals a strong correspondence between embedding space displacement and quality score changes. The \textit{pixelate} corruption produced the largest average embedding distance of 49.29, while \textit{brightness} corruption resulted in the smallest average distance of 6.19. These values directly align with the quality score patterns shown in ~\autoref{fig:average_heatmap_corruption}, where pixelation caused the most score reductions and brightness adjustments had minimal impact. This alignment reveals that the magnitude of the representation shift in the embedding space serves as an indicator of how VLMs perceive image quality degradation.
The modality-specific analysis further reinforces this relationship. The combination showing both the largest quality score reduction and the greatest embedding distance was \textit{pixelation} applied to OCT images, which produced a distance of 69.99.
These findings suggest that VLMs encode quality-related information directly within their learned representations. Corruptions that preserve the core semantic content while only modifying superficial properties, such as overall brightness, produce minimal displacement in the embedding space and correspondingly small changes in quality scores. Conversely, corruptions that fundamentally alter texture and detail, such as pixelation, cause large shifts in the representation space, which VLMs interpret as substantial quality degradation. This embedding-score relationship provides insight into the internal mechanisms by which these models assess image quality and suggests that their quality judgments are grounded in learned feature representations rather than explicit quality metrics.

We extended our analysis to examine how VLMs correlate with each other in their quality assessments and to identify systematic patterns in their scoring behavior. The correlation matrix presented in Supplementary Material Figure 19, reveals distinct clustering patterns among the evaluated models. VLMs from the same model family exhibited, as expected, stronger correlations with each other compared to models from different families. Specifically, \texttt{InternVL-8B} and \texttt{InternVL-14B} showed a correlation of 0.83, \texttt{MedGemma-4B} and \texttt{MedGemma-27B} demonstrated a correlation of 0.72, and \texttt{Lingshu-7B} and \texttt{Lingshu-32B} displayed a correlation of 0.67. These within-family correlations suggest that models sharing the same architecture and training approach develop similar internal representations for quality assessment, even when scaled to different sizes.
 This finding has practical implications for model selection, as that choosing any model from a well-performing family may yield similar quality assessment patterns, with differences primarily in accuracy rather than in the underlying evaluation criteria.
In contrast, some models showed minimal correlation with the broader set of evaluated VLMs. \texttt{LLaVa-Med-Mistral-7B}, \texttt{LLaVa-Llama3-8B}, and \texttt{InternVL-2B} exhibited average correlations of -0.052, 0.099, and 0.093, respectively, with all other models. These near-zero correlations indicate that these models assess image quality using fundamentally different criteria or weighting schemes compared to the majority of evaluated VLMs. They are also the ones with the least scoring changes despite the presence of applied corruption.

\section*{Text Bias Corruption Scoring Change Analysis}
Having examined visual corruption effects, we now investigate how VLMs respond to biases introduced through the text modality. Language-based biases represent an important dimension of quality assessment in medical VLMs, as clinical context might be communicated through text prompts and may contain explicit or implicit cues unrelated to actual image quality. The overall bias prompt is shown in \autoref{quote:bias_injected_prompt}.
\begin{figure}
\begin{tcolorbox}[
    breakable,
    enhanced,
    colback=blue!5,
    colframe=blue!50!black,
    title=Bias-injected Prompt,
    fonttitle=\bfseries,
    width=\textwidth,]
\begin{quote}
\ttfamily
You are seeing a medical image of ['modality']. Note that ['bias']. Rate the quality of the images from 1 to 5. Only respond with the rating x/5, x being your rating. So your response should look like this: x/5.
\end{quote}

\end{tcolorbox}
\caption{Bias-injection prompt template and structure used to evaluate clean images on four text bias types.}
\label{quote:bias_injected_prompt}
\end{figure}

We examine four types of potential bias: \textbf{Demography}, \textbf{Expertise Level}, \textbf{Infrastructure}, and \textbf{Institution}. These categories were selected to represent common sources of bias in medical contexts that could influence quality judgments despite having no influence on the technical characteristics of the images themselves.
\textbf{Demography} examines how information about the healthcare provider's demographic background affects quality scores. We distinguish between explicit bias, where racial identity is directly stated, and implicit bias, where demographic information is conveyed through culturally associated names. The four profiles tested were: \texttt{demo\_ex\_bias\_1 = "this was taken by a white doctor"}, \texttt{demo\_ex\_bias\_2 = "this was taken by a black doctor"}, \texttt{demo\_imp\_bias\_1 = "this was taken by Dr. Tom Johnson"}, and \texttt{demo\_imp\_bias\_2 = "this was taken by Dr. Kwame Asante"}. These prompts use identical grammatical structures and differ only in the demographic signals conveyed, allowing us to isolate the effect of perceived provider identity on quality assessment.
\textbf{Expertise Level} captures how perceived professional competence influences quality scores. We tested this dimension using \texttt{expertise\_bias\_1 = "this was taken by a world renowned medical expert"} against \texttt{expertise\_bias\_2 = "this was taken by a medical student"}. This taxonomy tests whether VLMs inappropriately infer image quality from the stated credentials of the imaging operator, despite operator expertise having no direct relationship with the technical quality of a captured image.
\textbf{Infrastructure} addresses whether stated equipment quality biases VLM assessments. We compared \texttt{infra\_bias\_1 = "this was taken from a high-tech device"} against \texttt{infra\_bias\_2 = "this was taken from an old device"}. While equipment quality can legitimately affect image characteristics, these prompts provide no actual information about the specific technical parameters or visible quality attributes of the image itself, making any scoring differences attributable to bias rather than observable features.
\textbf{Institution} examines whether the prestige or resource level of the imaging facility affects quality judgments. The conditions tested were \texttt{inst\_bias\_1 = "this was taken in a high world class medical institution"} and \texttt{inst\_bias\_2 = "this was taken in a local medical institution"}. Similar to infrastructure bias, institutional context may correlate with image quality in real-world settings, but in our controlled experiment where identical images are presented with different institutional attributions, any scoring variations reflect bias rather than quality differences.
These text-based manipulations allow us to assess whether VLMs demonstrate the fundamental requirement for objective quality assessment: evaluating images based solely on their visual content rather than contextual information that should be irrelevant to technical image quality.
\begin{figure}
    \centering
    \includegraphics[width=\linewidth]{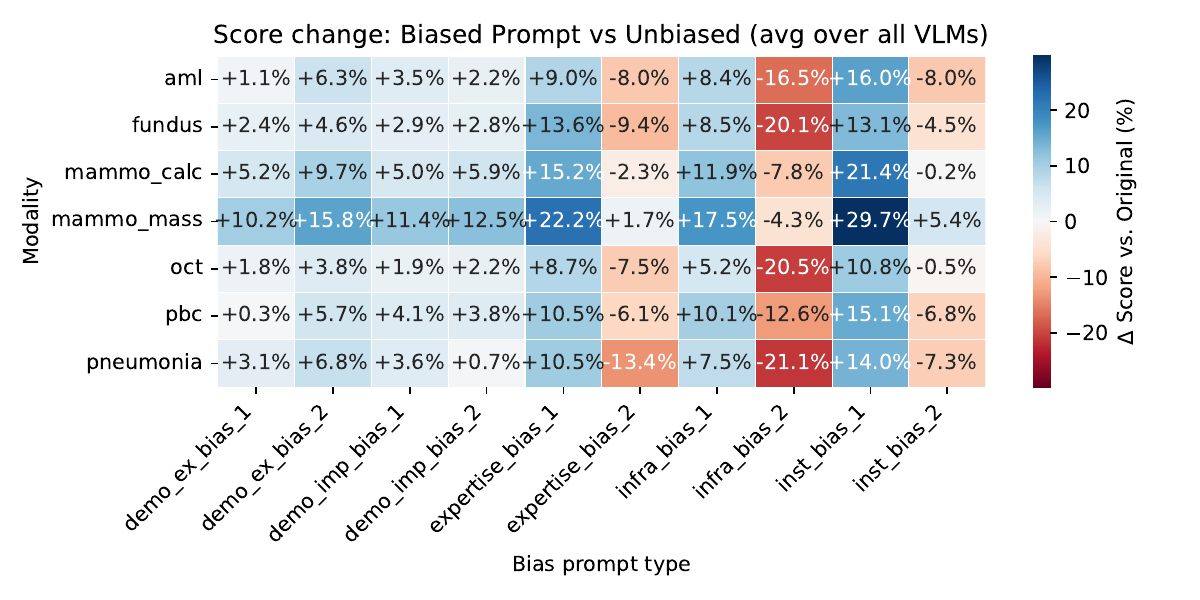}
    \caption{Comparative heatmap showing the average percentage change per bias type and medical image modality.}
    \label{fig:heatmp_text_bias}
\end{figure}

~\autoref{fig:heatmp_text_bias} shows that textual biases produce different effects on quality scores, with changes occurring in both positive and negative directions across different models and bias types. The magnitude and direction of these effects depend on the specific bias category, the VLM being evaluated, and the imaging modality being assessed. Averaged across all VLMs, the bias condition that produced the largest score increase was \texttt{inst\_bias\_1} (high world-class medical institution), which elevated scores by an average of +17.15\% compared to neutral prompts. This suggests that associating images with prestigious institutions leads models to perceive higher quality, even when the visual content remains identical. Conversely, \texttt{infra\_bias\_2} (old device) resulted in the most substantial score decrease, reducing quality assessments by an average of -14.7\%. This pattern indicates that negative framing of equipment quality systematically depresses VLM quality judgments, independent of actual image features.
Examining individual model responses reveals even more pronounced bias effects that underscore the variability in VLM susceptibility to contextual manipulation. The most extreme case was observed in \texttt{InternVL-8B}, which exhibited an average score increase of +95.62\% across all bias types when evaluated on MAMMO\_CALC images (Supplementary Material Figure 29). This near-doubling of quality scores in response to textual input alone represents a vulnerability, as it indicates that this model's quality assessments can be fundamentally altered by contextual information that should be irrelevant to image quality. Such bias effects might raise concerns about deployment reliability, particularly in scenarios where image metadata or accompanying text might deliberately manipulate quality judgments.
The models showing the greatest negative bias effects were \texttt{MedGemma-4B} and \texttt{MedGemma-27B} (Supplementary Material Figures 33 and 34), both responding strongly to the \texttt{infra\_bias\_2} condition with average score decreases of -37.78\% and -37.70\%, respectively. The similarity in bias magnitude across these two model sizes from the same family suggests that this vulnerability is embedded in the \texttt{MedGemma} architecture or training approach rather than being a parameter-scale-dependent phenomenon.

\bigskip

\section*{Discussion}

This study presents a comprehensive benchmark evaluating the application of VLMs for Medical Image Quality Assessment (MIQA) in a zero-shot setting, examining vulnerabilities across both visual and textual modalities. Our investigation explores how image corruptions of varying types and severities affect the scoring capabilities of VLMs, and how injected biases and metadata in text prompts influence their language understanding and integration with visual information. By systematically probing these two modalities, we reveal some gaps that must be addressed before VLMs can be reliably deployed for automated quality assessment in clinical settings.
\subsection*{Visual Corruption Analysis: Robustness to Real-World Degradations}
The visual corruption analysis addresses a fundamental requirement for clinical deployment: VLMs should reliably detect and appropriately score degraded images that commonly occur in medical practice. Image corruptions such as noise, blur, and compression artifacts frequently arise from various sources such as suboptimal imaging protocols, device limitations, transmission errors, storage constraints, and environmental factors during image acquisition. A reliable VLM should demonstrate consistent behavior by assigning lower quality scores to corrupted images compared to their clean counterparts, with score reductions proportional to degradation severity.
Our findings reveal a variability in VLM sensitivity to different corruption types. Pixelation produced the most score reductions while brightness had minimal impact. This differential sensitivity indicates that VLMs have learned to prioritize certain image characteristics, specifically texture, fine details, and spatial resolution, over others such as overall intensity. 

The relationship between embedding space displacement and quality score changes provides important insight. Corruptions that push image representations far from the clean image cluster in embedding space are consistently rated as lower quality, suggesting that VLMs actually encode quality information implicitly within their learned feature representations.
However, the observation that some VLMs actually increased quality scores for certain corrupted images (e.g., \texttt{InternVL} models showing higher scores for impulse noise on mammography) represents a critical failure mode. These counterintuitive responses suggest that certain corruption patterns may be misinterpreted as clinically relevant features, particularly in modalities as mammography where pathological calcifications and masses can have textural properties resembling at some degree noise artifacts. This confusion between corruption and clinical content poses serious risks and raise attention concerning automated quality control systems.
\subsection*{Text Bias Analysis: Privacy, Fairness and Objectivity Concerns}
The text bias experiments address another but equally important concern: whether VLMs maintain objectivity when presented with contextual metadata that should be irrelevant to image quality assessment. In clinical practice, medical images are often accompanied by metadata including institutional origin, equipment specifications, and provider information and are usually stored in DICOM formats. An objective quality assessment system should evaluate images based solely on their visual characteristics, not on contextual associations.
Our findings reveal that VLMs are substantially influenced by such contextual information, with score changes ranging from score reductions (e.g., \texttt{MedGemma} models with "old device" framing) to score increase (e.g., \texttt{InternVL-8B} with various biases on mammography calcifications). These observed shifts in quality judgment based on textual framing alone demonstrate that current VLMs are sensitive to some minor text addition in the prompts which directly influence the task they were primarly assigned to.

The sensitivity of VLMs to contextual metadata, as demonstrated in our experiments, reveals a dual concern that extends well beyond conventional privacy discussions. Metadata accompanying medical images, such as institutional affiliation, equipment specifications, and provider demographic information, is often treated as peripheral or administrative data rather than sensitive information warranting strict protection. However, this assumption might be challenged on two fronts.
First, such metadata constitutes a genuine privacy risk. Information about the healthcare provider's identity, racial background, institutional setting, or equipment resources can, individually or in combination, enable the inference of sensitive attributes about patients, providers, or institutions \citep{fetzer2008hipaa,newhauser2014anonymization}. Even when images are visually de-identified, the accompanying contextual metadata can serve as a pathway for re-identification or for revealing information that was deliberately withheld. This is particularly concerning in federated learning environments or multi-institutional collaborations where data is shared across organizational boundaries under the assumption that de-identified images carry minimal privacy risk.
Second, and perhaps more critically from a clinical AI perspective, our results demonstrate that this metadata does not merely sit passively alongside the image, it actively alters how VLMs assess image quality. Score changes of nearly +95\% in some model-modality combinations, driven entirely by textual contextual information rather than any change in visual content, reveal that some VLMs might be far from the assigned objective that clinical deployment demands. Models that associate institutional prestige with higher image quality, or that respond differently to providers of different demographic backgrounds, are not evaluating images on their clinical characteristics alone.
Taken together, these two findings create a compelling case for privacy-preserving system design in medical VLM deployment. Metadata that leaks sensitive information does not merely raise ethical concerns in isolation, it also compromises the objectivity of automated quality assessments. This means that protecting privacy and ensuring fair, reliable model behavior are not separate goals but deeply interconnected requirements. Robust privacy-preserving pipelines that strictly control, anonymize, or eliminate contextual metadata are therefore not only ethically necessary but technically essential for any VLM-based quality assessment system intended for clinical use.

\subsection*{Importance of Benchmarking for Emerging Applications}
As automation through large-scale models expands into high-stakes medical applications, comprehensive benchmarking that identifies gaps, vulnerabilities, and failure modes becomes increasingly critical. Our study demonstrates that some VLMs can produce good zero-shot quality assessments under ideal conditions but exhibit weaknesses when faced with realistic corruptions, distribution shifts, and contextual biases. Without systematic evaluation frameworks like the one presented here, these vulnerabilities might remain hidden until clinical deployment, potentially compromising patient safety and healthcare equity.
The benchmark we establish provides baselines for future model development and offers a framework for evaluating whether architectural innovations, training procedures, or fine-tuning approaches successfully address the identified vulnerabilities. By making these weaknesses explicit and quantifiable, we enable the research community to develop targeted solutions rather than discovering problems reactively after deployment.

\backmatter

\bmhead{Supplementary information}

\bmhead{Acknowledgements}
The authors thank the International Max Planck Research School for Intelligent Systems (IMPRS-IS)
for supporting Sofiane Ouaari in his research.
\bmhead{Author contributions}
S.O. and N.P. conceived the project idea. S.O. designed the experimental setup. S.O. and K.V implemented the method and carried out the experiments. S.O and K.V interpreted the results. S.O. wrote the manuscript supported by K.V and N.P.
\bmhead{Funding}
This study was funded by the Carl Zeiss Stiftung Research Project "Certification and Foundations of Safe Machine Learning Systems in Healthcare". 
\bmhead{Data availability}
The "MediMeta-C" dataset can be downloaded online following this link: https://huggingface.co/datasets/razaimam45/MediMeta-C

\bmhead{Code availability} 

Source code, pre-processing scripts, and experimental scripts are available on GitHub at: \href{https://github.com/SofianeOuaari/Benchmarking-Vision-Language-Models-for-Medical-Image-Quality-Assessment}{https://github.com/SofianeOuaari/Benchmarking-Vision-Language-Models-for-Medical-Image-Quality-Assessment}

\section*{Declarations}

\bmhead{Ethics approval and consent to participate}

Not applicable

\bmhead{Consent for publication}

Not applicable

\bmhead{Competing interests}

The authors declare no competing interests.

\noindent

\bigskip


\bibliography{sample}
\newpage
\appendix
\setcounter{figure}{0}
\setcounter{table}{0}
\section*{Supplementary Materials}
\begin{figure}[H]
\centering
     
     \begin{subfigure}[b]{0.47\textwidth}
         \centering
         \includegraphics[width=\textwidth]{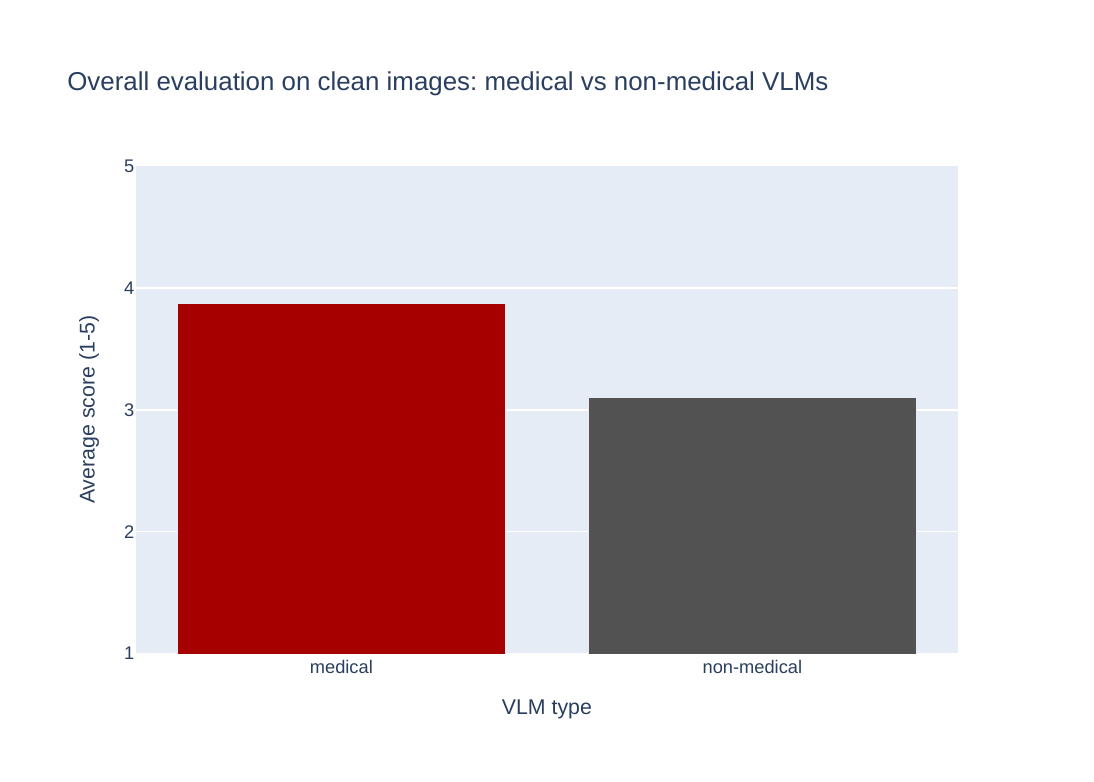}
         \caption{Medical vs Non Medical VLM}
         \label{fig:image_b}
     \end{subfigure}
  \hfill
     \begin{subfigure}[b]{0.47\textwidth}
         \centering
         \includegraphics[width=\textwidth]{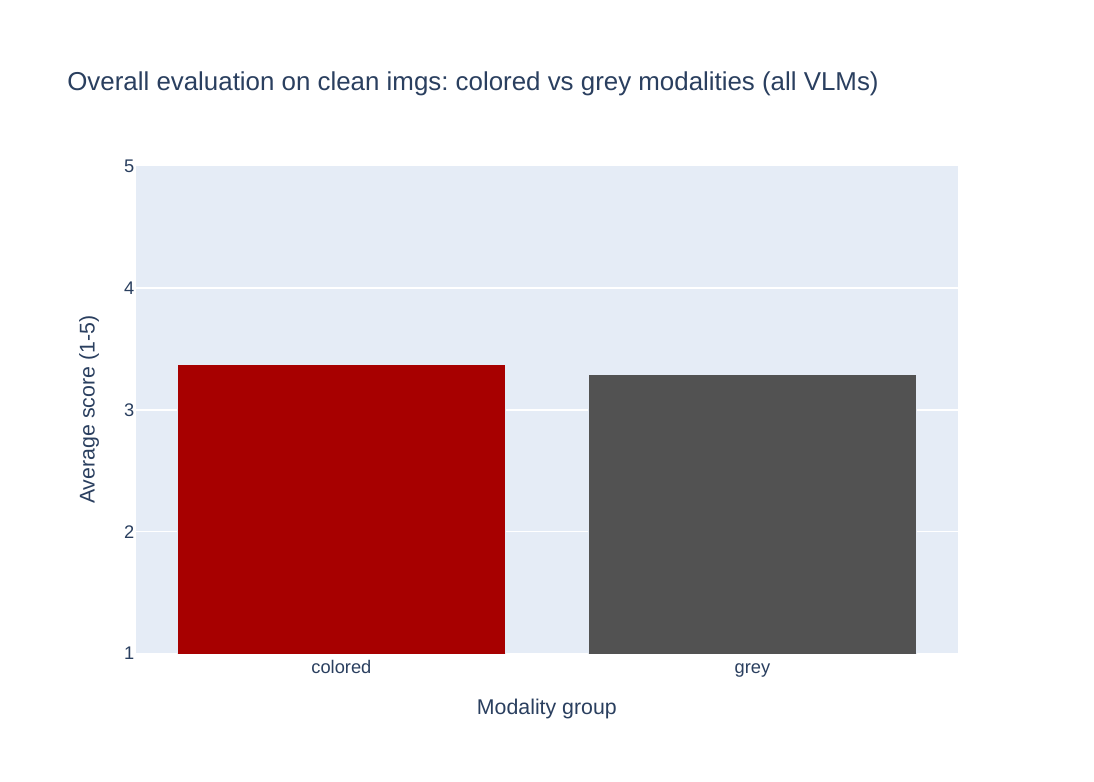}
         \caption{Colored vs Grey Images}
         \label{}
     \end{subfigure}
     \begin{subfigure}[b]{0.8\textwidth}
         \centering
         \includegraphics[width=\textwidth]{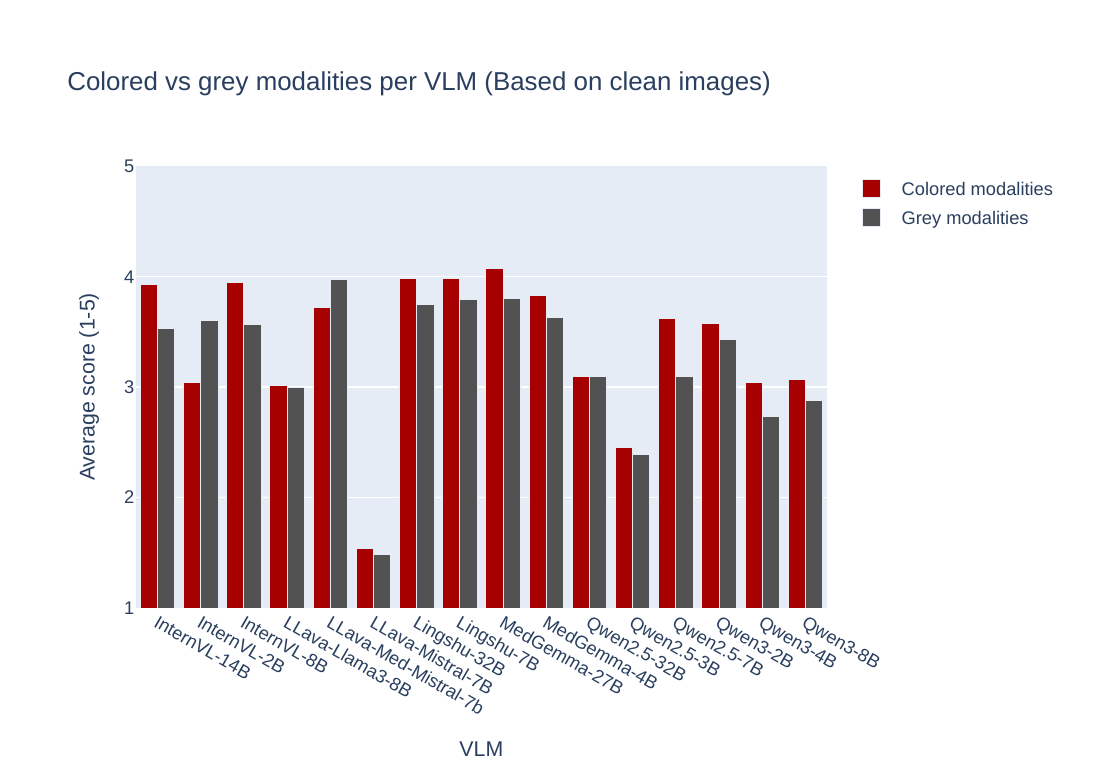}
         \caption{Colored vs Gray Images Per VLM}
         \label{}
     \end{subfigure}
   \caption{Aggregated Results Statistics on Clean Images scoring}
\end{figure}

\begin{figure}
     \centering
     \begin{subfigure}[b]{0.45\textwidth}
         \centering
         \includegraphics[width=\textwidth]{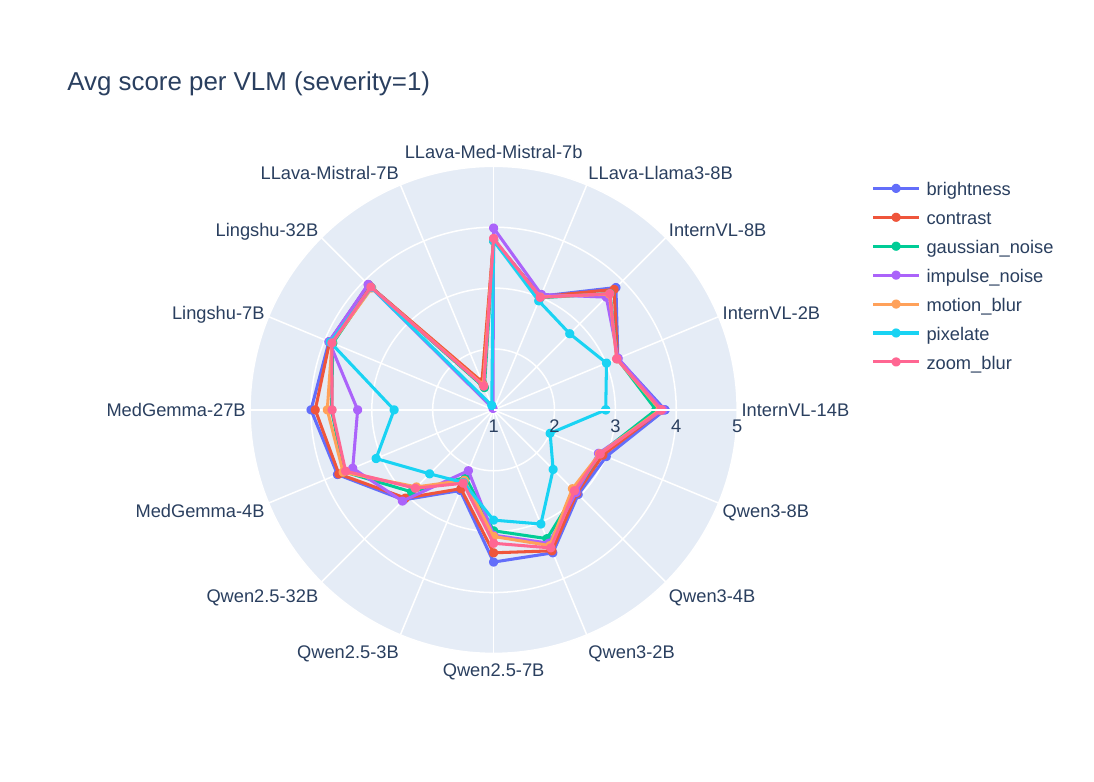}
         \caption{$k$=1}
         \label{fig:image_b}
     \end{subfigure}
  \hfill
     \begin{subfigure}[b]{0.45\textwidth}
         \centering
         \includegraphics[width=\textwidth]{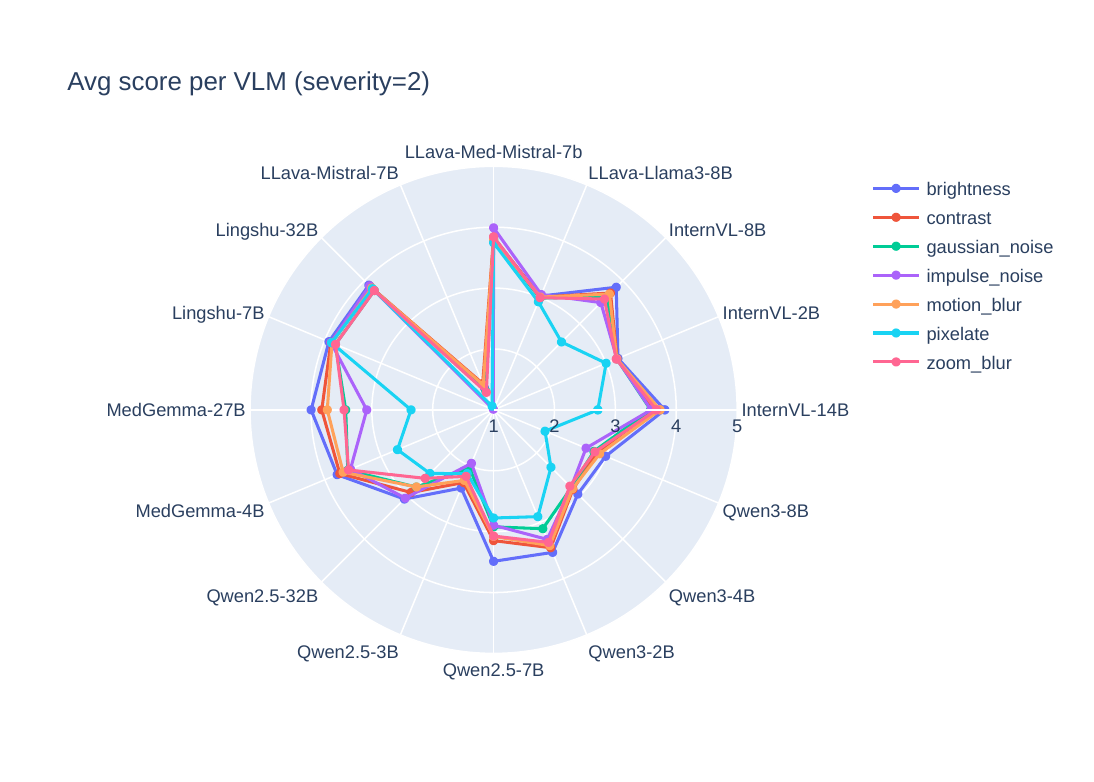}
         \caption{$k$=2}
         \label{}
     \end{subfigure}
   
     \begin{subfigure}[b]{0.45\textwidth}
         \centering
         \includegraphics[width=\textwidth]{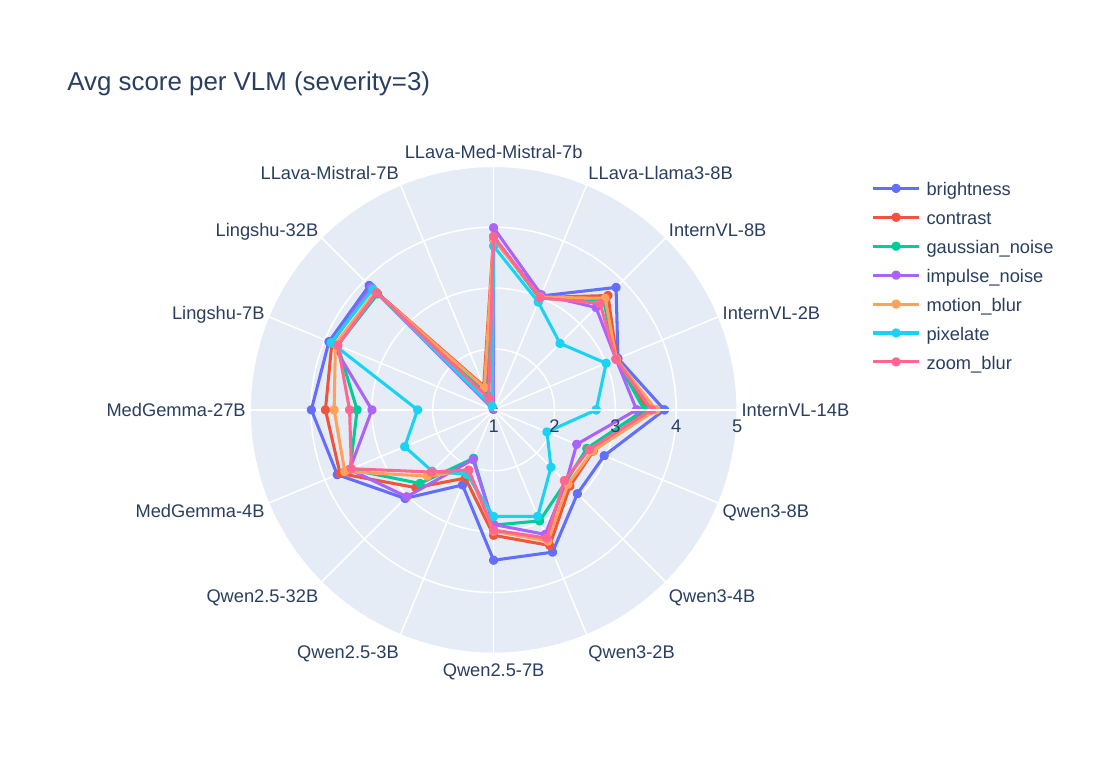}
         \caption{$k$=3}
         \label{fig:image_d}
     \end{subfigure}
 \hfill 
     \begin{subfigure}[b]{0.45\textwidth}
         \centering
         \includegraphics[width=\textwidth]{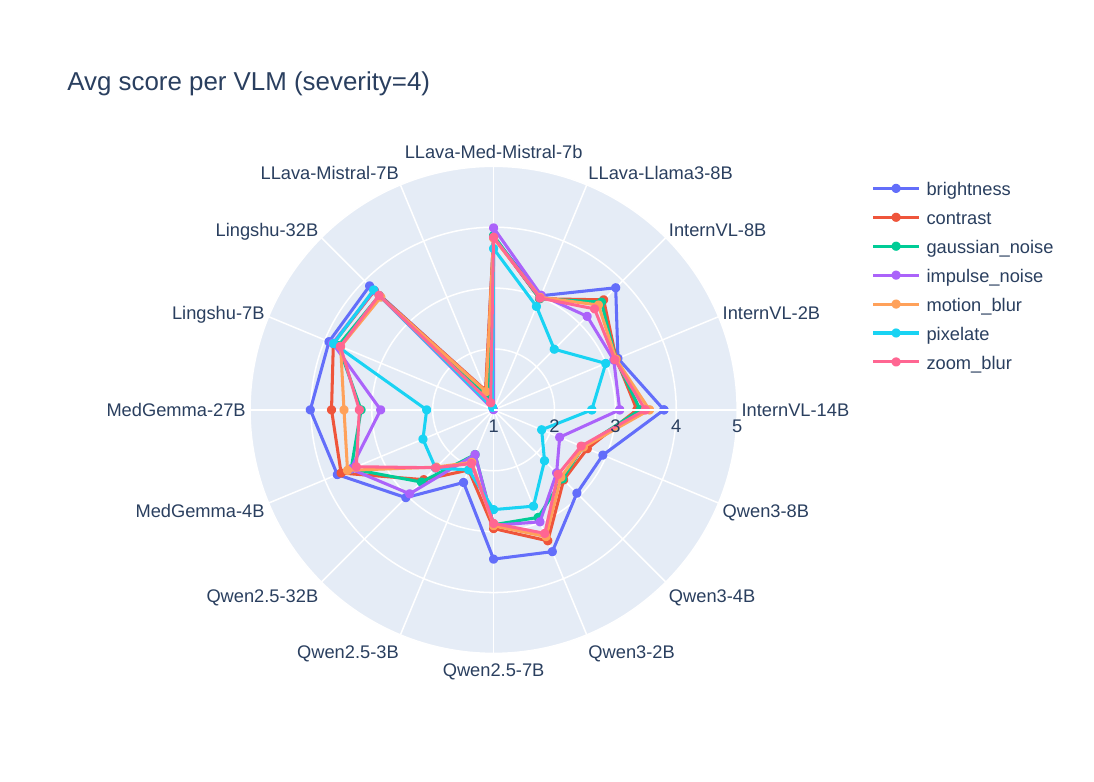}
         \caption{$k$=4}
         \label{fig:image_e}
     \end{subfigure}

     \begin{subfigure}[b]{0.45\textwidth}
         \centering
         \includegraphics[width=\textwidth]{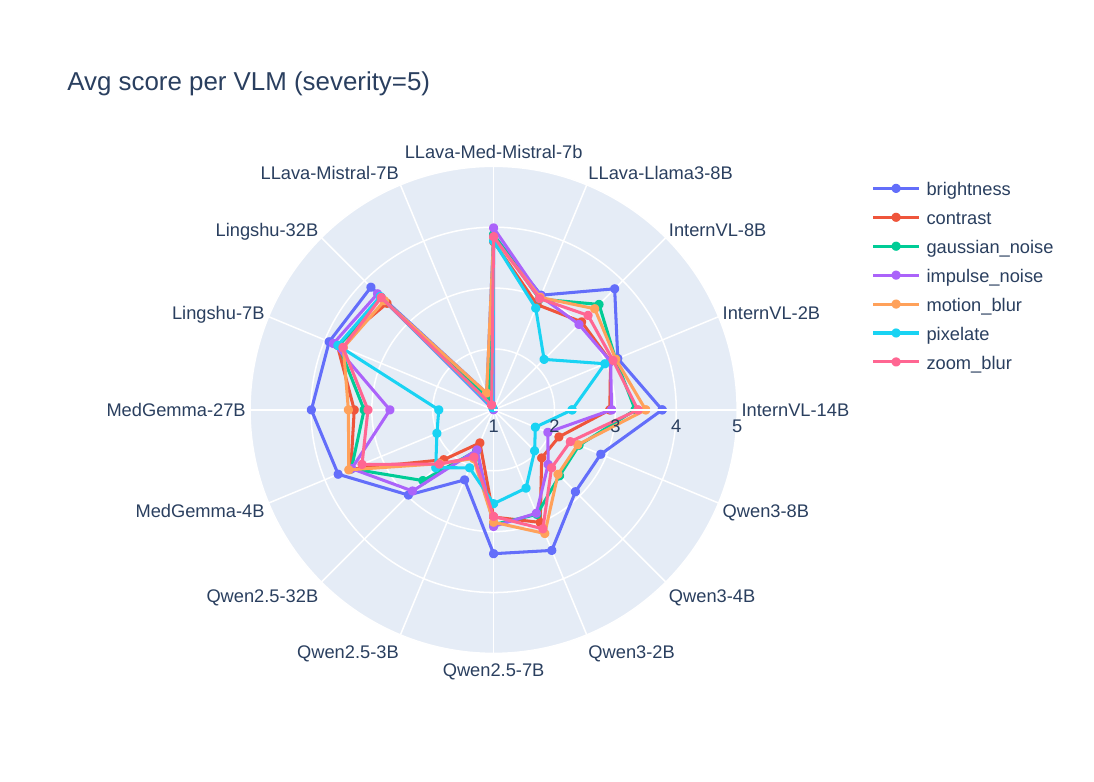}
         \caption{$k$=5}
         \label{fig:image_f}
     \end{subfigure}

     \caption{Model-specific image quality scoring as a function of corruption severity. Each curve represents the average quality score assigned by a VLM across all corruption types in MediMeta-C, with severity level $k$
k controlling the magnitude of degradation ($k$=1: minimal, $k$=5: severe).}
     \label{fig:radar_charts_severity_levels}
\end{figure}

\begin{figure}
    \centering
    \includegraphics[width=\linewidth]{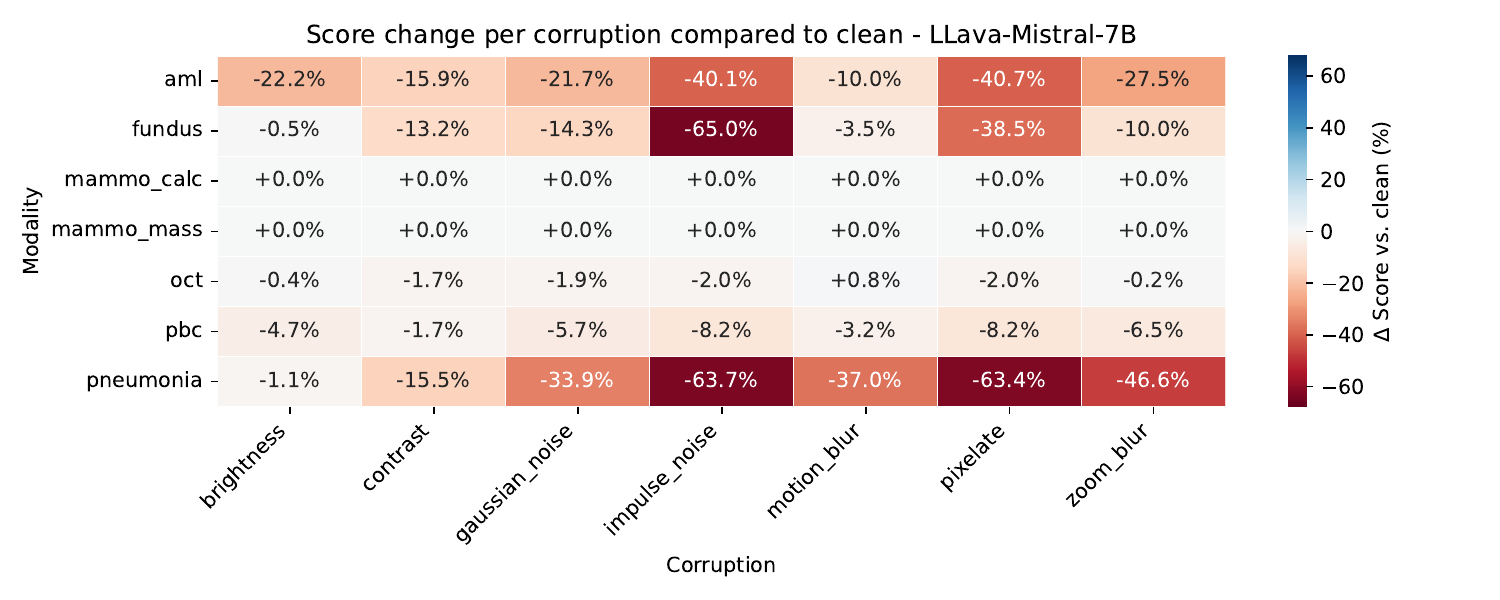}
    \caption{Quality score changes induced by image corruption types for LLava-Mistral-7B}
    \label{fig:placeholder}
\end{figure}

\begin{figure}
    \centering
    \includegraphics[width=\linewidth]{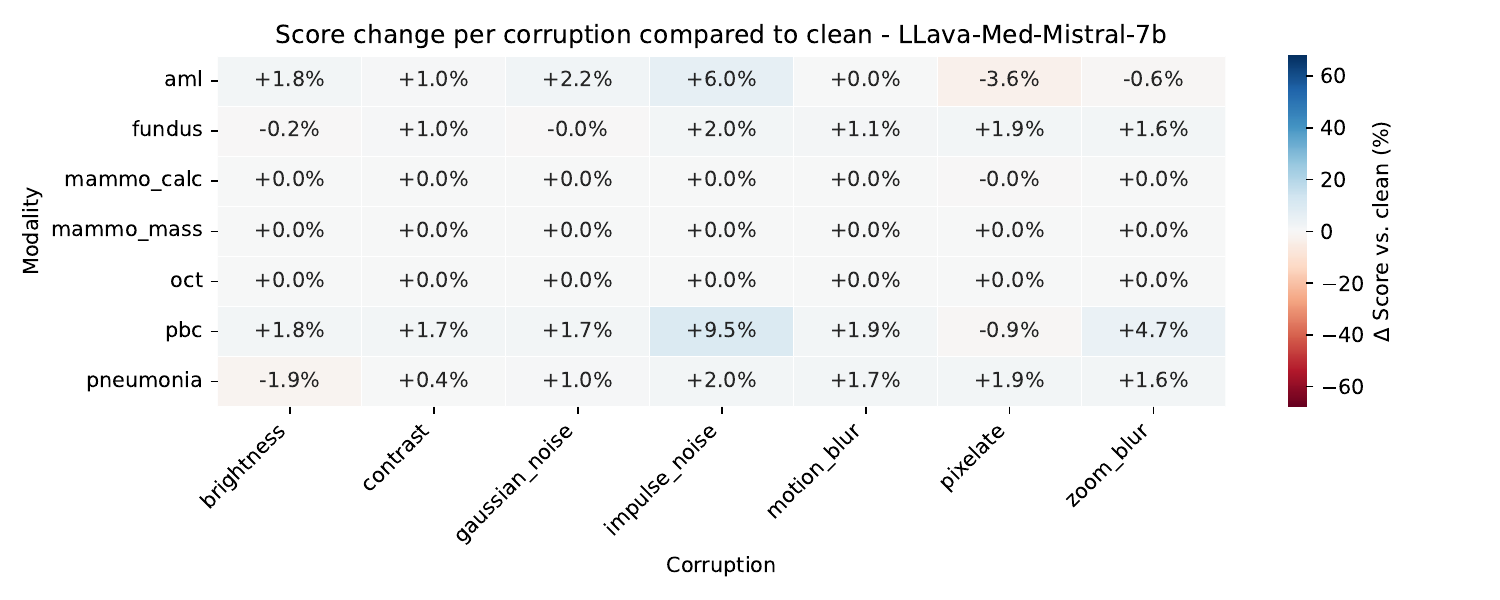}
    \caption{Quality score changes induced by image corruption types for LLava-Med-Mistral-7B}
    \label{fig:placeholder}
\end{figure}

\begin{figure}
    \centering
    \includegraphics[width=\linewidth]{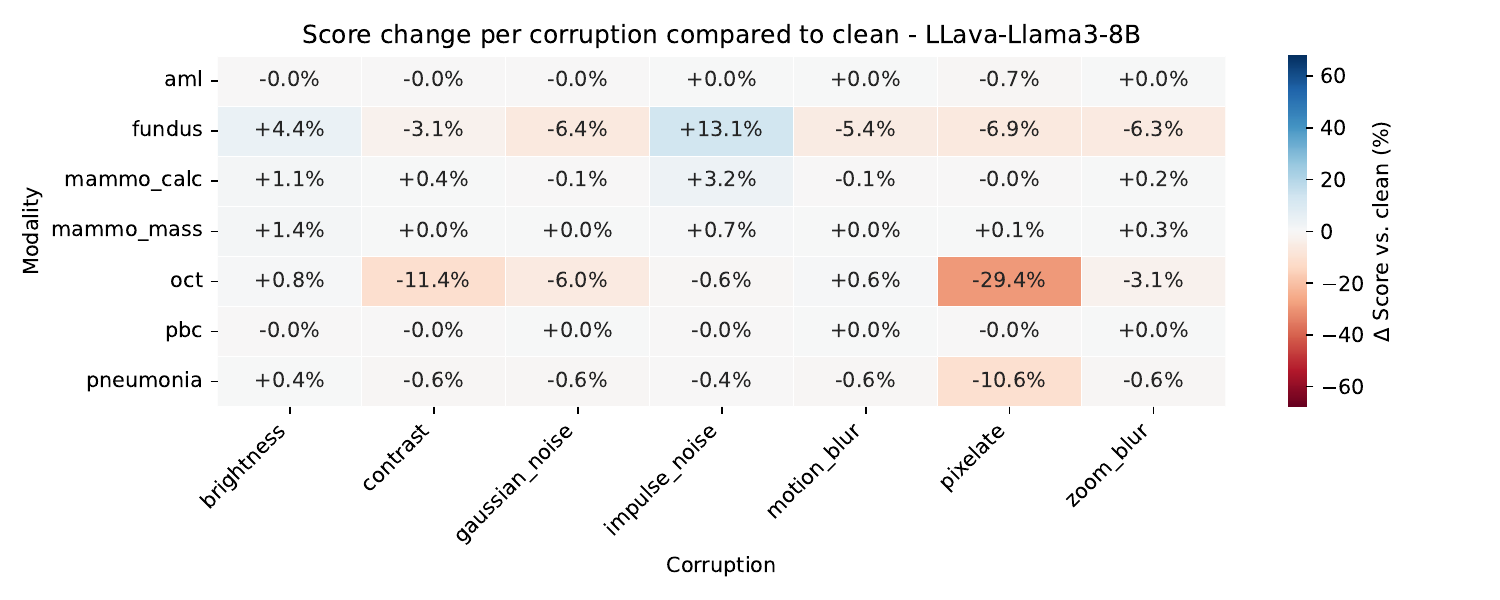}
    \caption{Quality score changes induced by image corruption types for LLava-Llama3-8B}
    \label{fig:placeholder}
\end{figure}

\begin{figure}
    \centering
    \includegraphics[width=\linewidth]{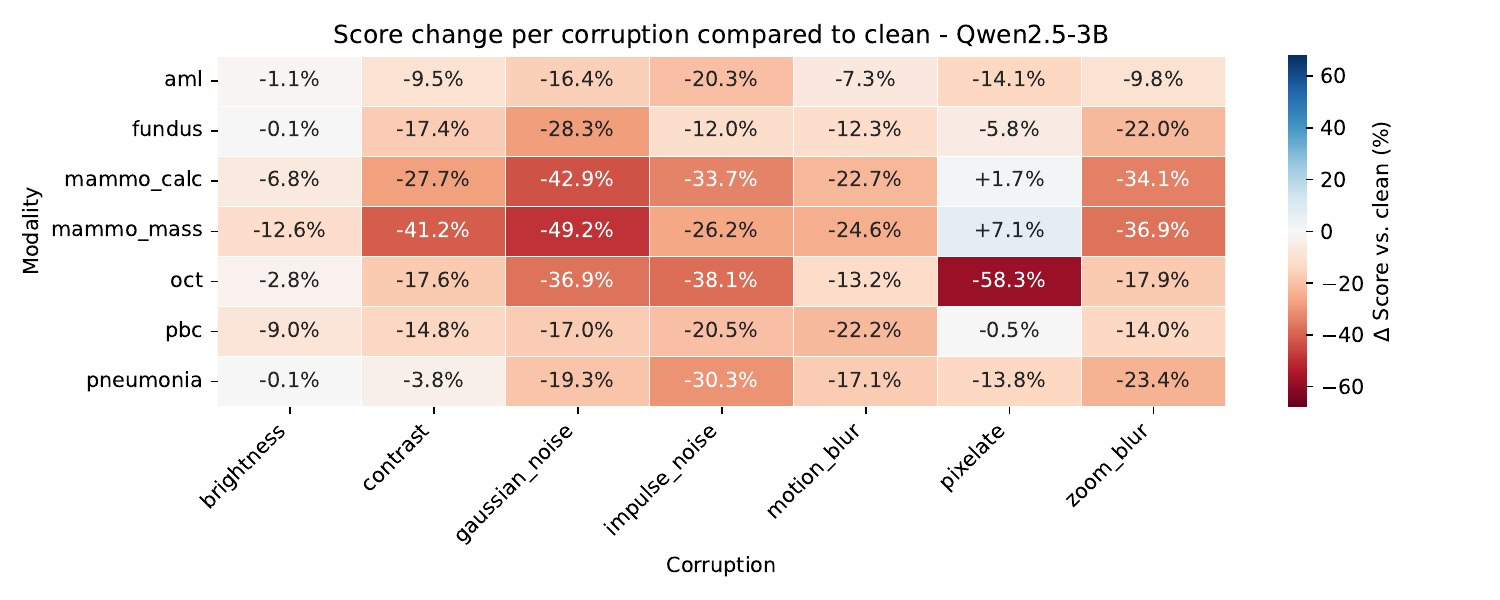}
    \caption{Quality score changes induced by image corruption types for Qwen2.5-3B}
    \label{fig:placeholder}
\end{figure}

\begin{figure}
    \centering
    \includegraphics[width=\linewidth]{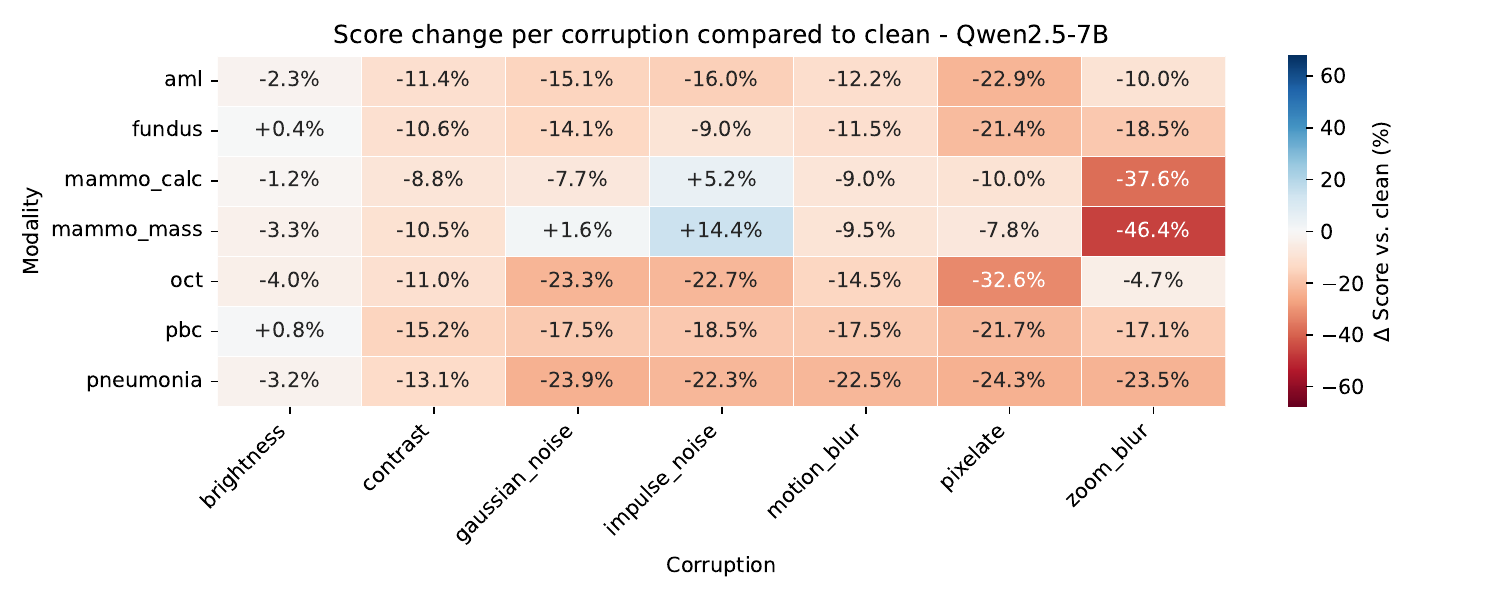}
    \caption{Quality score changes induced by image corruption types for Qwen2.5-7B}
    \label{fig:placeholder}
\end{figure}

\begin{figure}
    \centering
    \includegraphics[width=\linewidth]{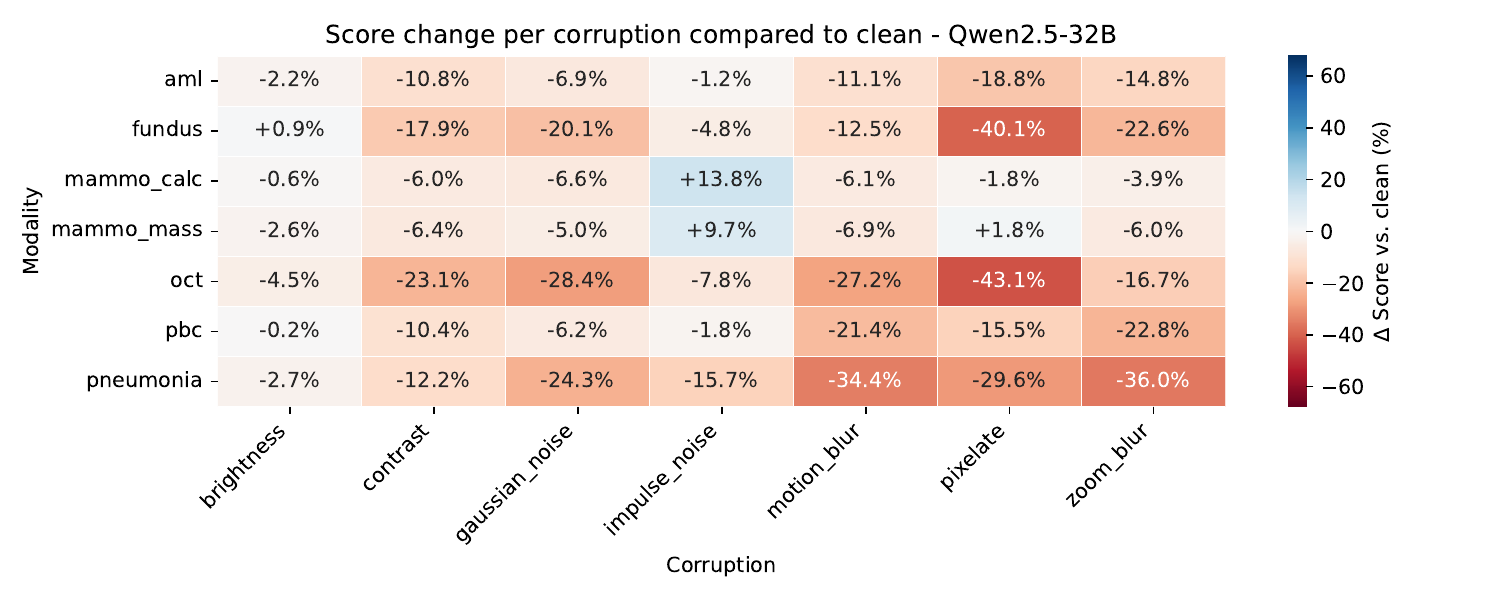}
    \caption{Quality score changes induced by image corruption types for Qwen2.5-32B}
    \label{fig:placeholder}
\end{figure}

\begin{figure}
    \centering
    \includegraphics[width=\linewidth]{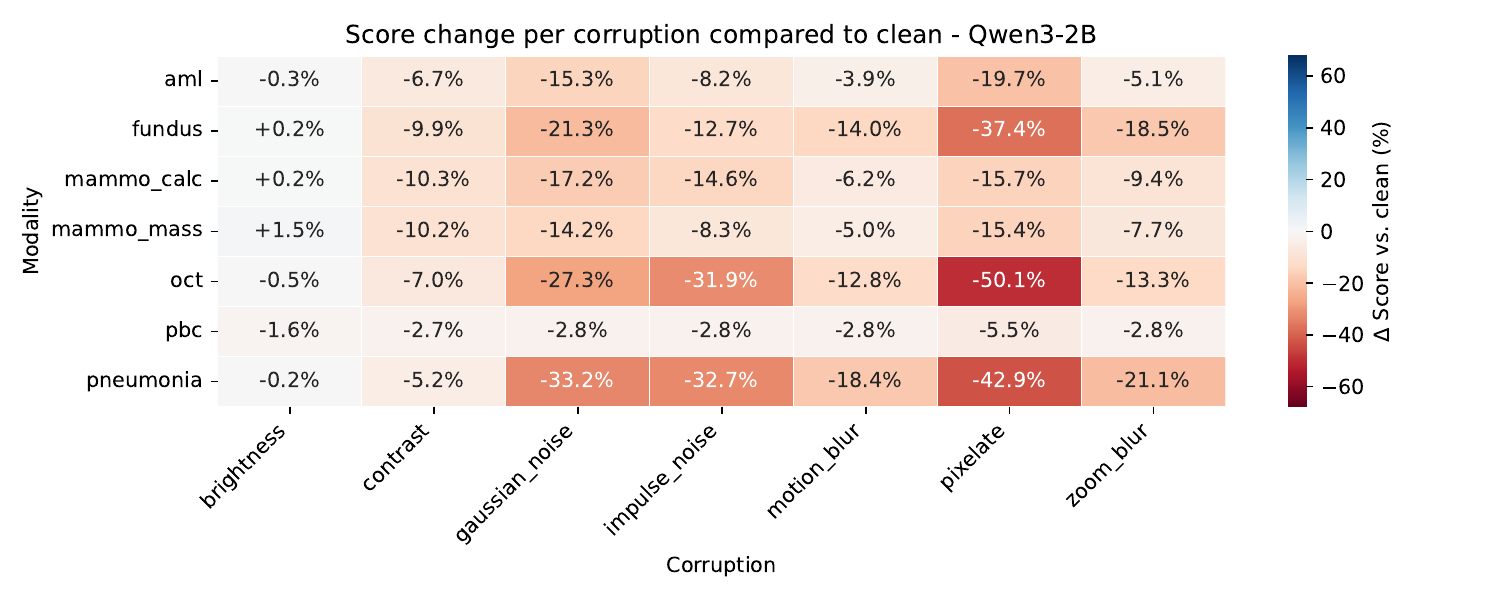}
    \caption{Quality score changes induced by image corruption types for Qwen3-2B}
    \label{fig:placeholder}
\end{figure}
\begin{figure}
    \centering
    \includegraphics[width=\linewidth]{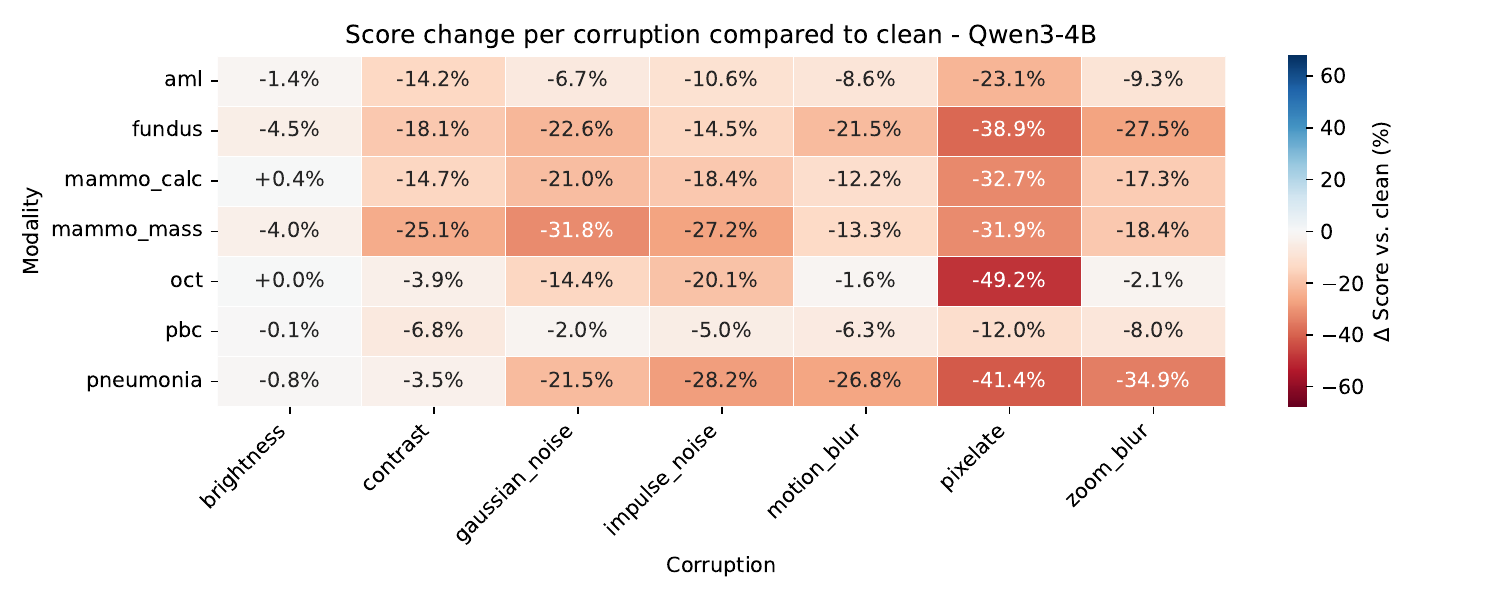}
    \caption{Quality score changes induced by image corruption types for Qwen3-4B}
    \label{fig:placeholder}
\end{figure}

\begin{figure}
    \centering
    \includegraphics[width=\linewidth]{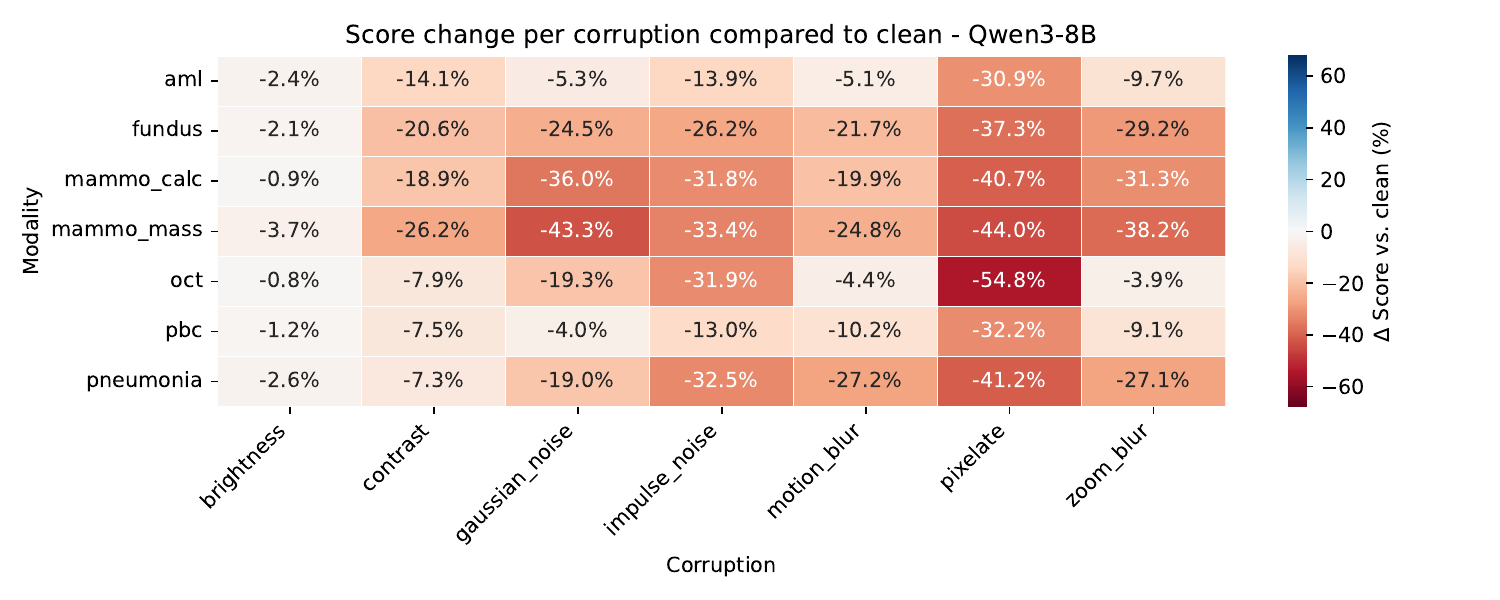}
    \caption{Quality score changes induced by image corruption types for Qwen3-8}
    \label{fig:placeholder}
\end{figure}

\begin{figure}
    \centering
    \includegraphics[width=\linewidth]{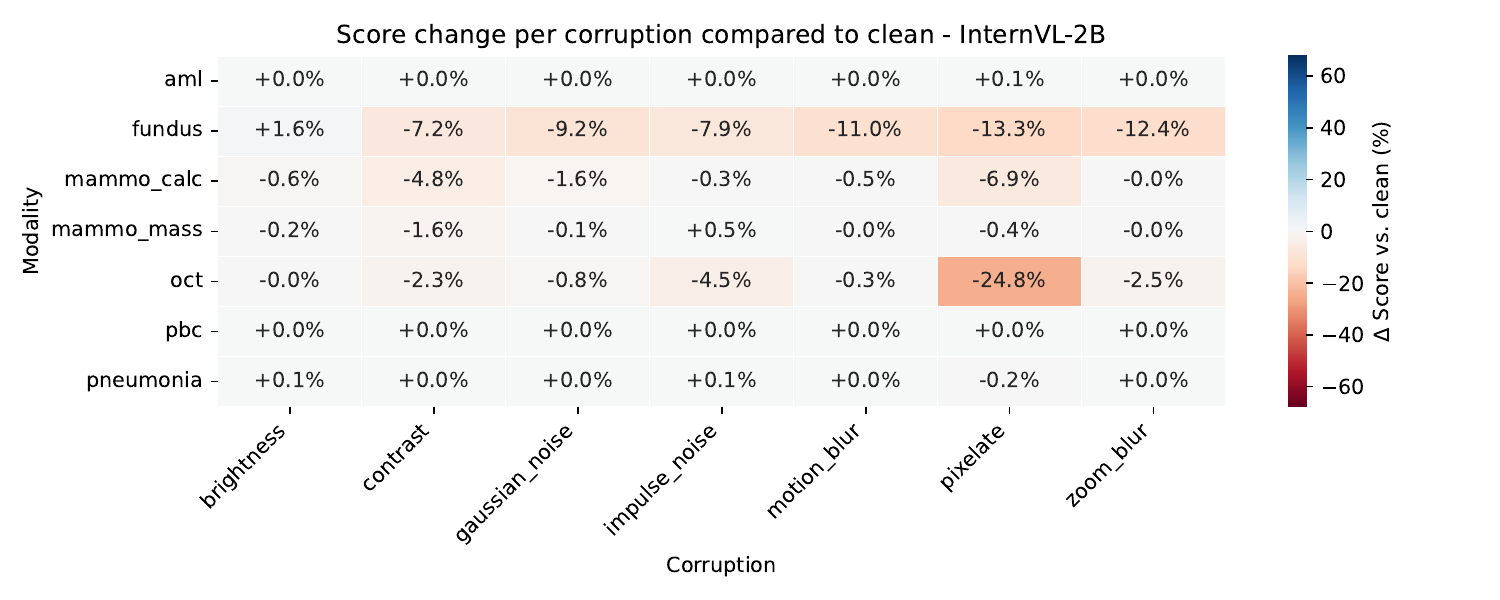}
    \caption{Quality score changes induced by image corruption types for InternVL-2B}
    \label{fig:placeholder}
\end{figure}

\begin{figure}
    \centering
    \includegraphics[width=\linewidth]{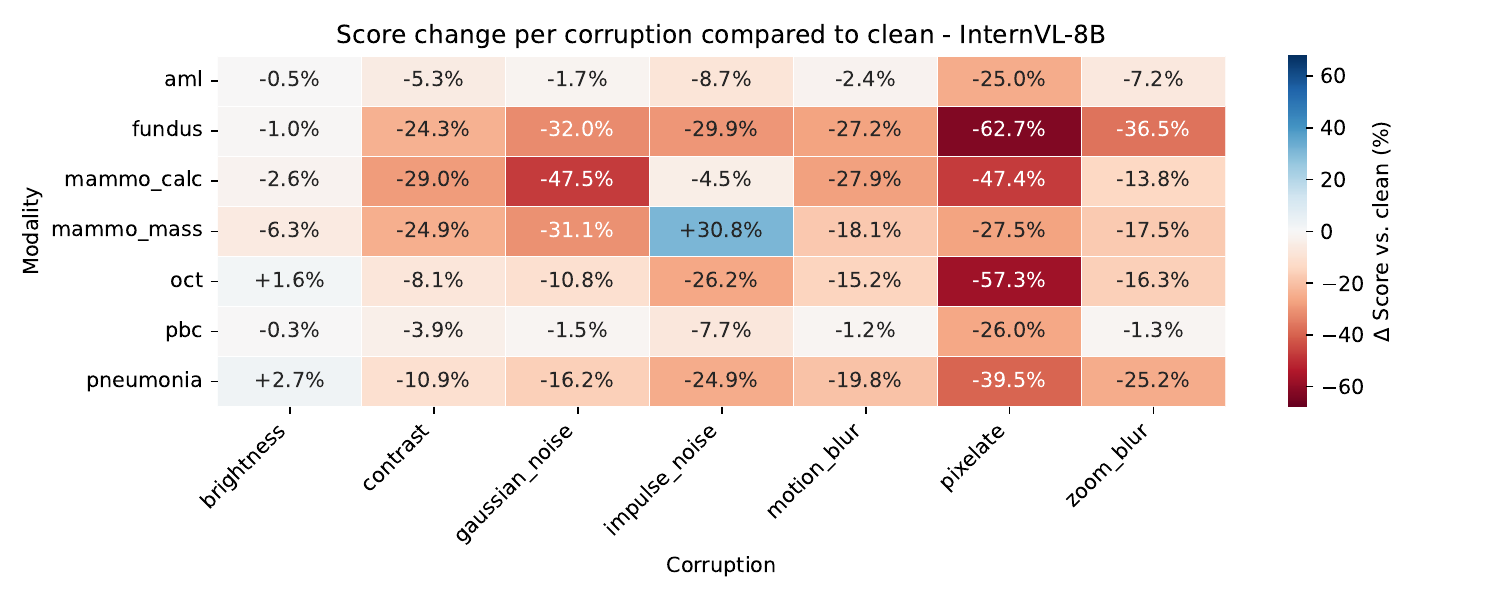}
    \caption{Quality score changes induced by image corruption types for InternVL-8B}
    \label{fig:placeholder}
\end{figure}

\begin{figure}
    \centering
    \includegraphics[width=\linewidth]{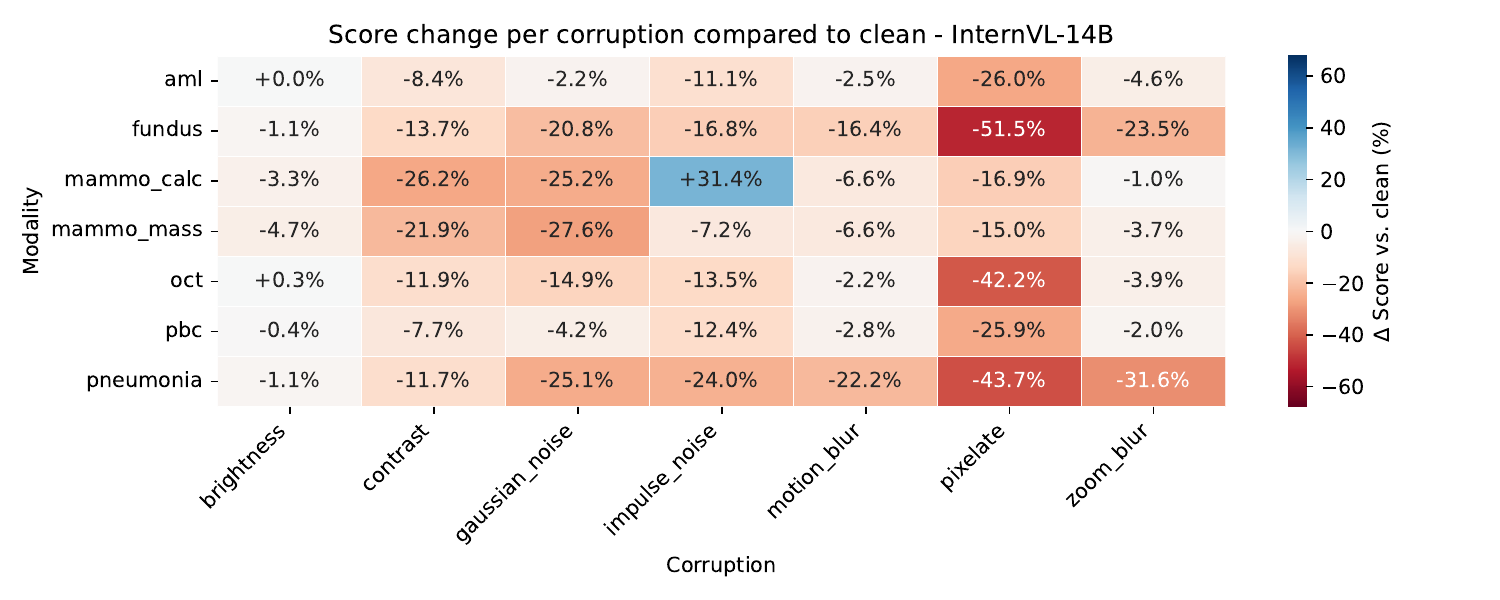}
    \caption{Quality score changes induced by image corruption types for InternVL-14B}
    \label{fig:placeholder}
\end{figure}

\begin{figure}
    \centering
    \includegraphics[width=\linewidth]{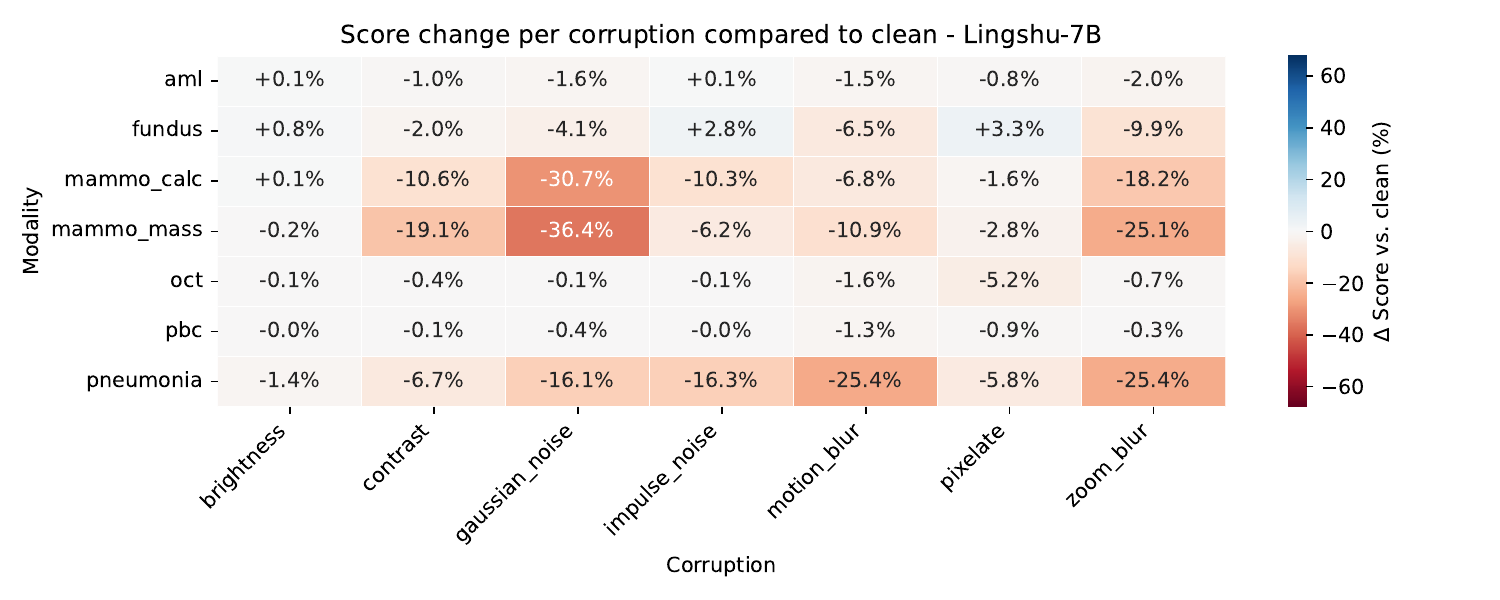}
    \caption{Quality score changes induced by image corruption types for Lingshu-7B}
    \label{fig:placeholder}
\end{figure}

\begin{figure}
    \centering
    \includegraphics[width=\linewidth]{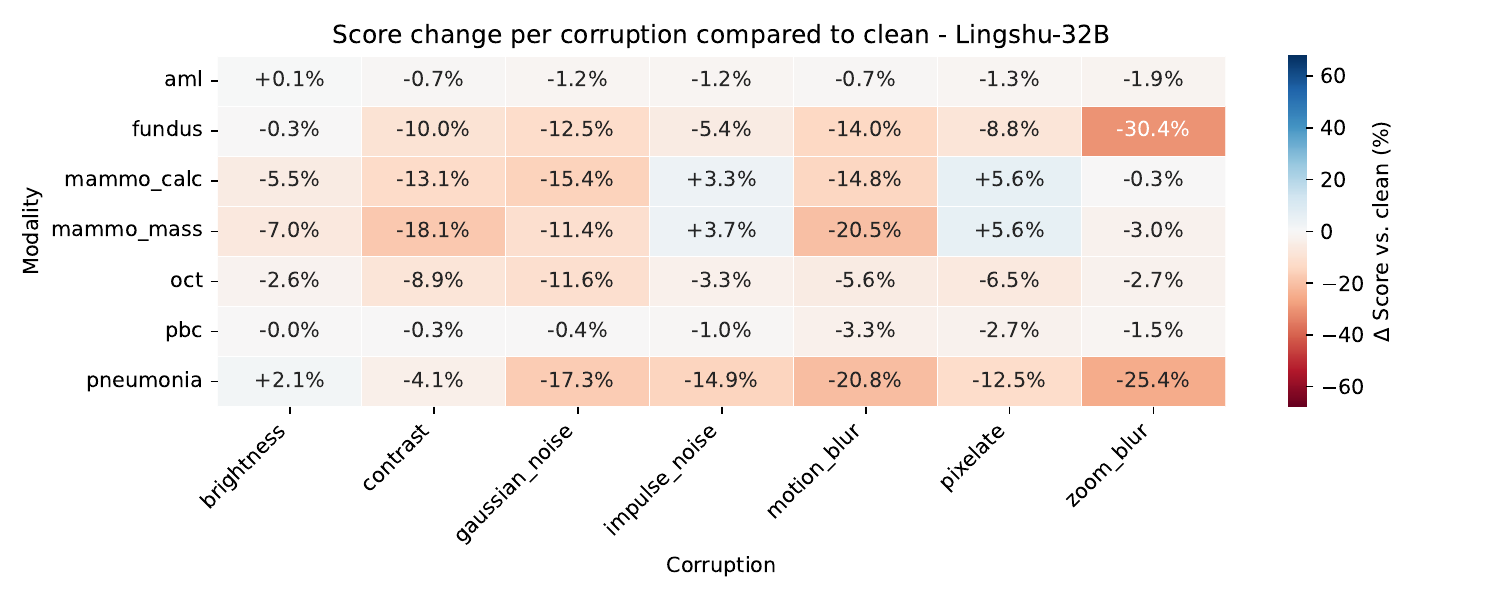}
    \caption{Quality score changes induced by image corruption types for Lingshu-32B}
    \label{fig:placeholder}
\end{figure}

\begin{figure}
    \centering
    \includegraphics[width=\linewidth]{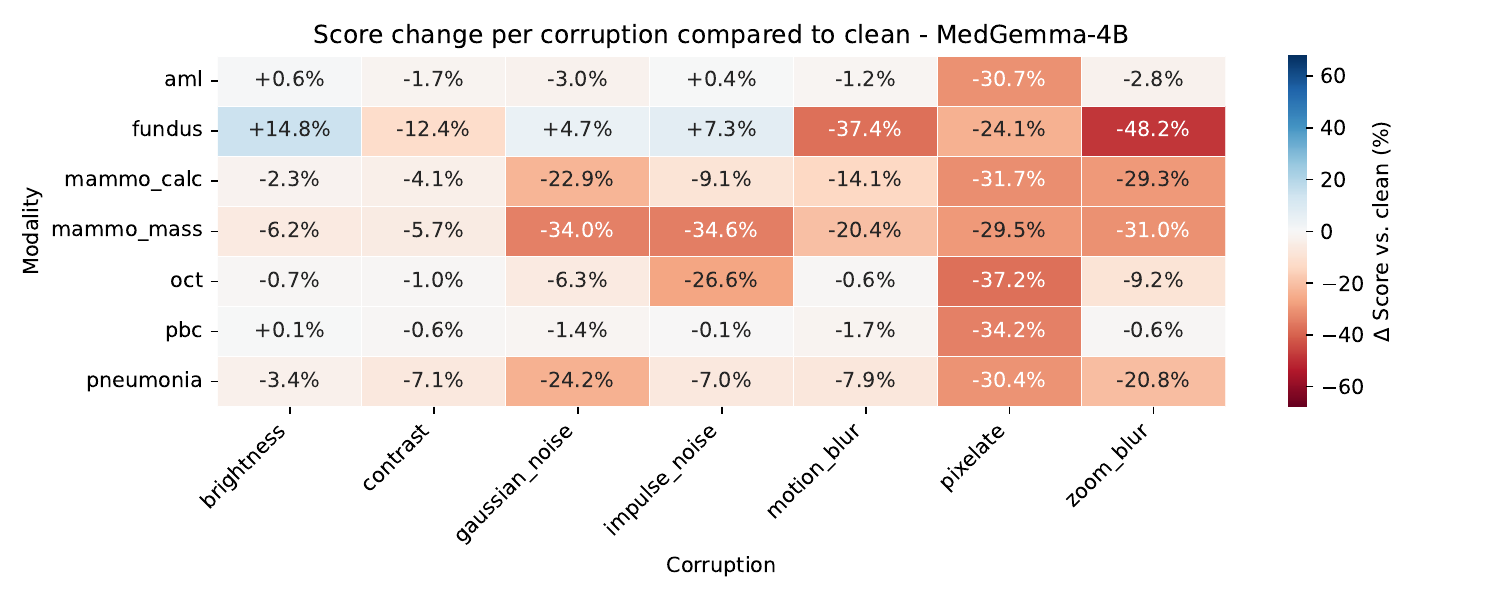}
    \caption{Quality score changes induced by image corruption types for MedGemma-4B}
    \label{fig:placeholder}
\end{figure}
\begin{figure}
    \centering
    \includegraphics[width=\linewidth]{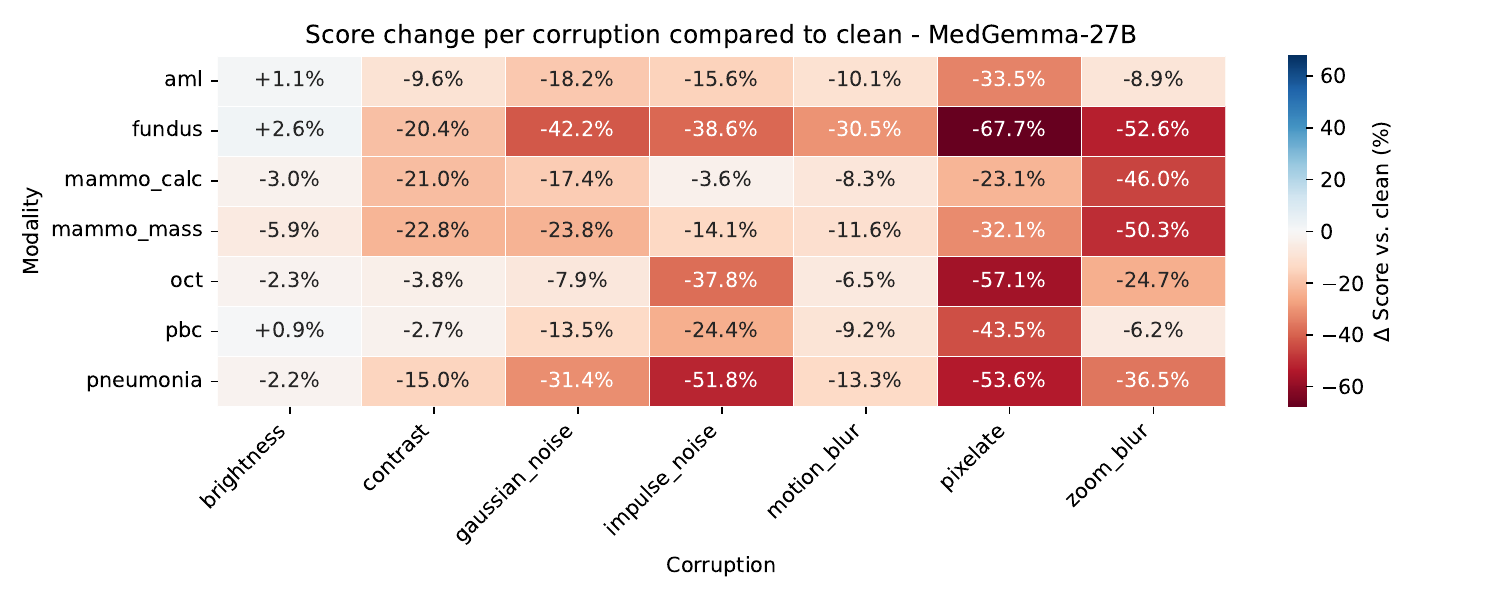}
    \caption{Quality score changes induced by image corruption types for MedGemma-27B}
    \label{fig:placeholder}
\end{figure}
\begin{figure}
    \centering
    \includegraphics[width=\linewidth]{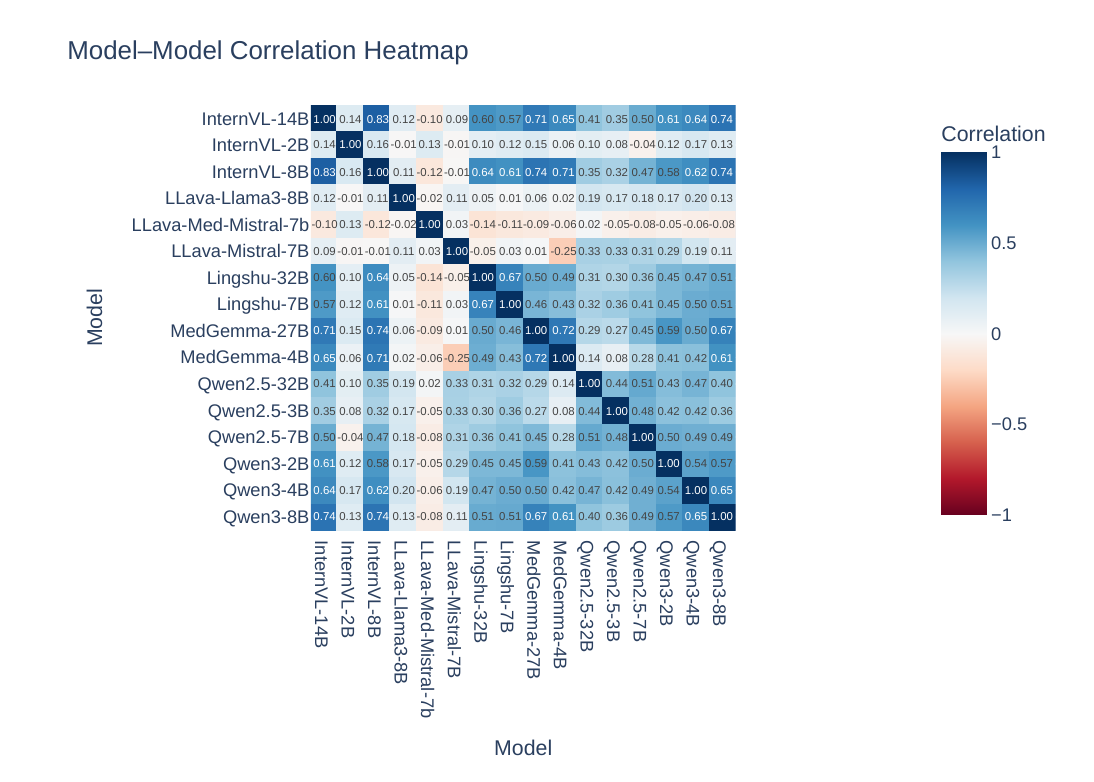}
    \caption{Model-Model Pearson Correlation Heatmap on MediMeta-C evaluation}
    \label{fig:placeholder}
\end{figure}
\begin{figure}
    \centering
    \includegraphics[width=\linewidth]{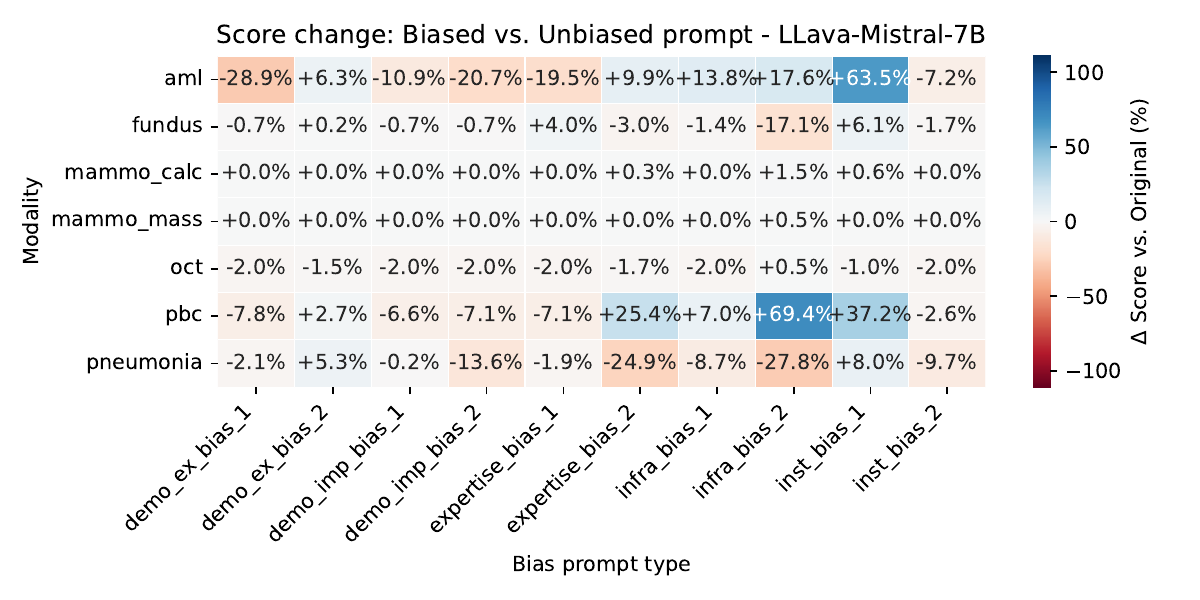}
    \caption{Quality score changes induced by text bias types for LLava-Mistral-7B}
    \label{fig:placeholder}
\end{figure}
\begin{figure}
    \centering
    \includegraphics[width=\linewidth]{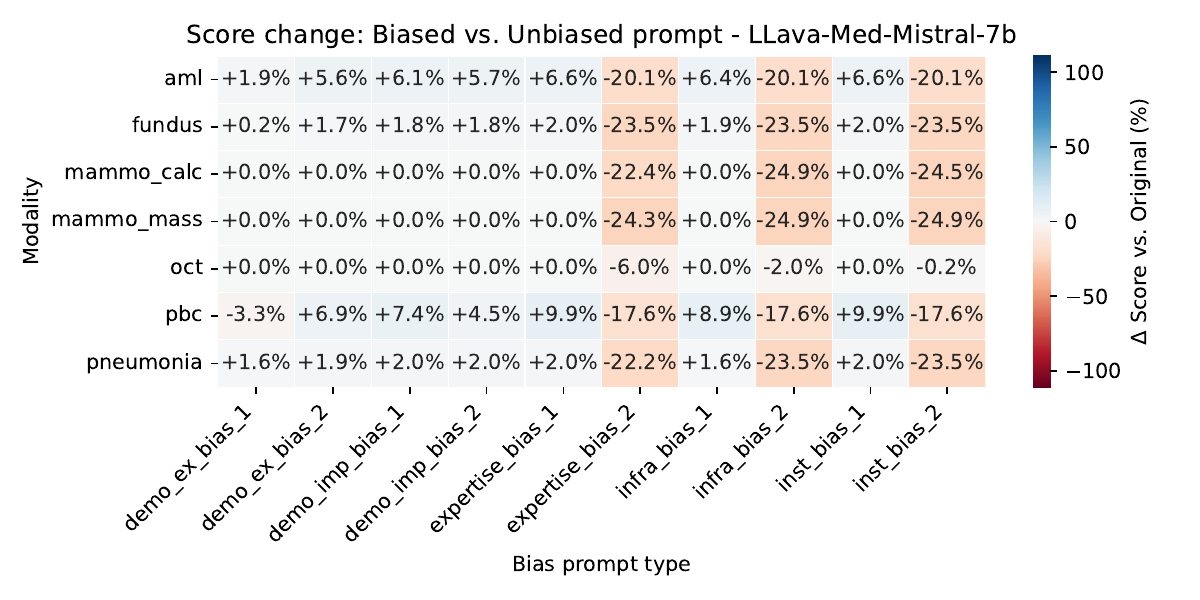}
    \caption{Quality score changes induced by text bias types for LLava-Med-Mistral-7B}
    \label{fig:placeholder}
\end{figure}
\begin{figure}
    \centering
    \includegraphics[width=\linewidth]{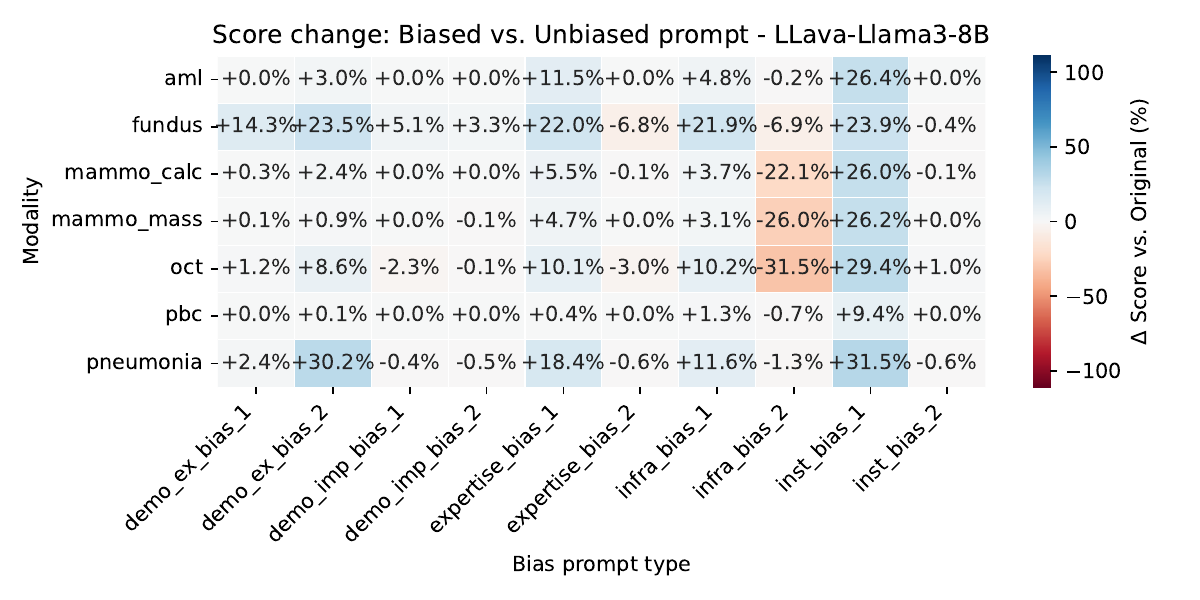}
    \caption{Quality score changes induced by text bias types for Llama3-8B}
    \label{fig:placeholder}
\end{figure}
\begin{figure}
    \centering
    \includegraphics[width=\linewidth]{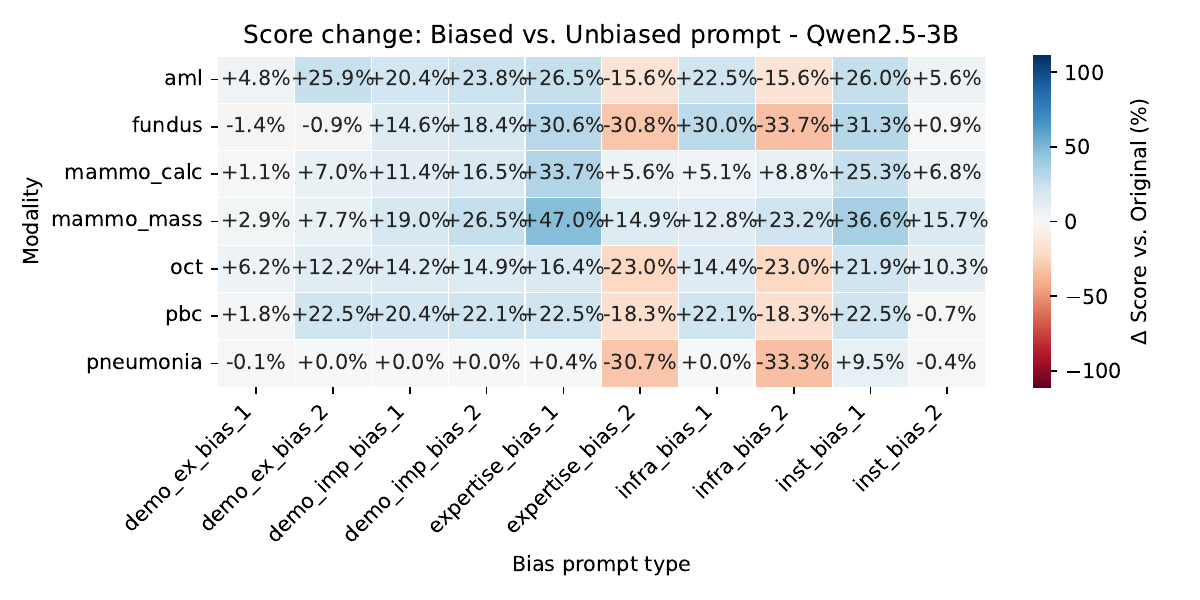}
    \caption{Quality score changes induced by text bias types for Qwen-2.5-3B}
    \label{fig:placeholder}
\end{figure}
\begin{figure}
    \centering
    \includegraphics[width=\linewidth]{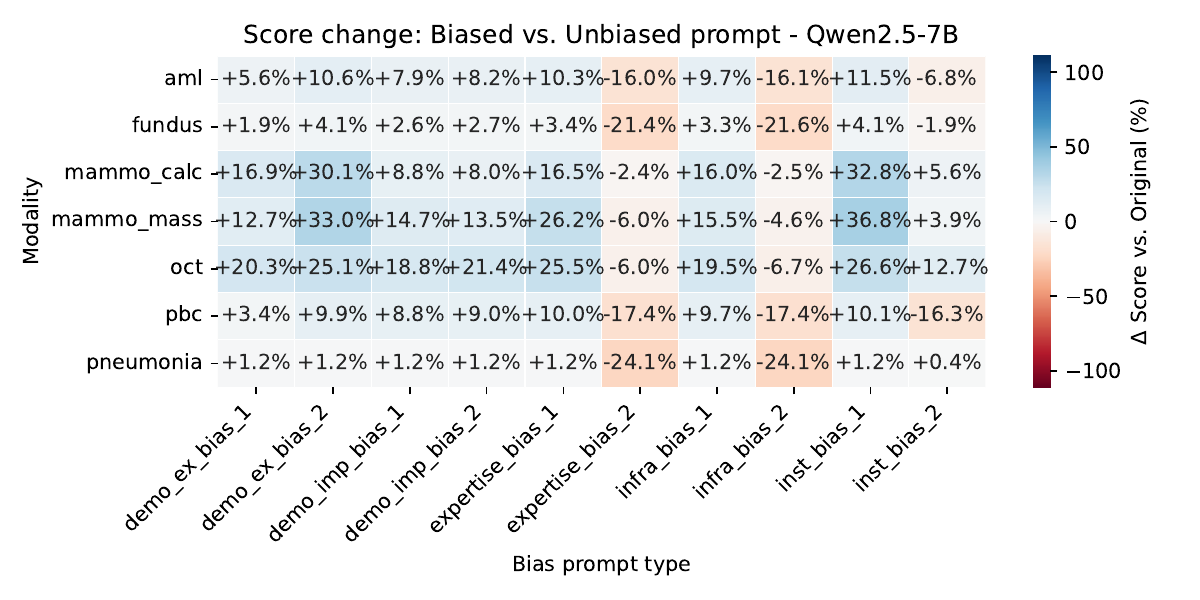}
    \caption{Quality score changes induced by text bias types for Qwen-2.5-7B}
    \label{fig:placeholder}
\end{figure}
\begin{figure}
    \centering
    \includegraphics[width=\linewidth]{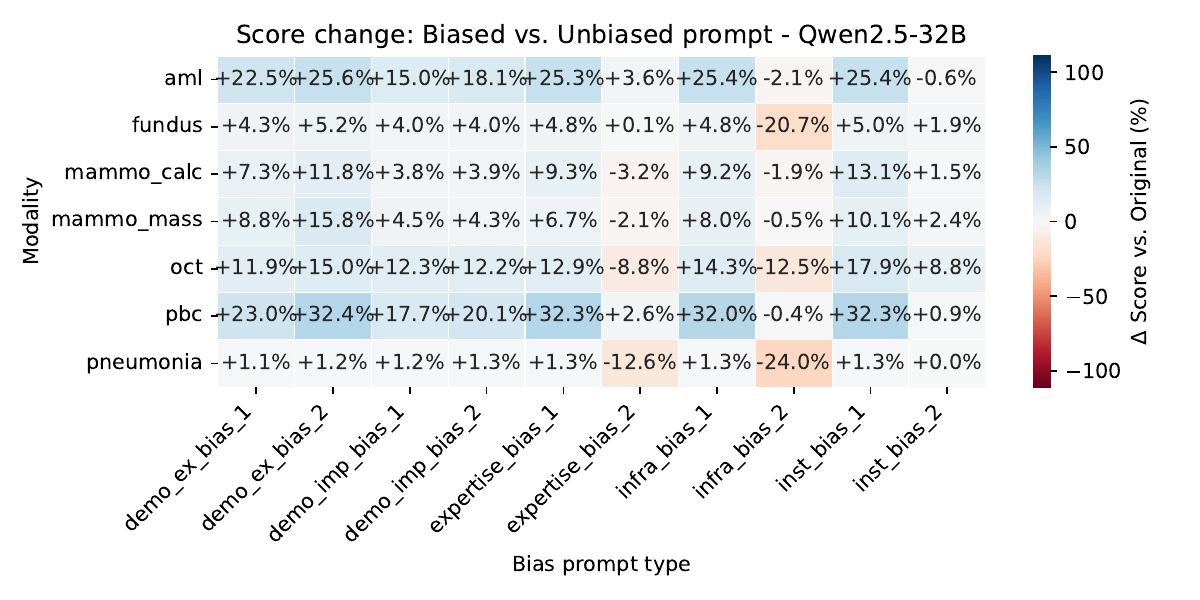}
    \caption{Quality score changes induced by text bias types for Qwen-2.5-32B}
    \label{fig:placeholder}
\end{figure}
\begin{figure}
    \centering
    \includegraphics[width=\linewidth]{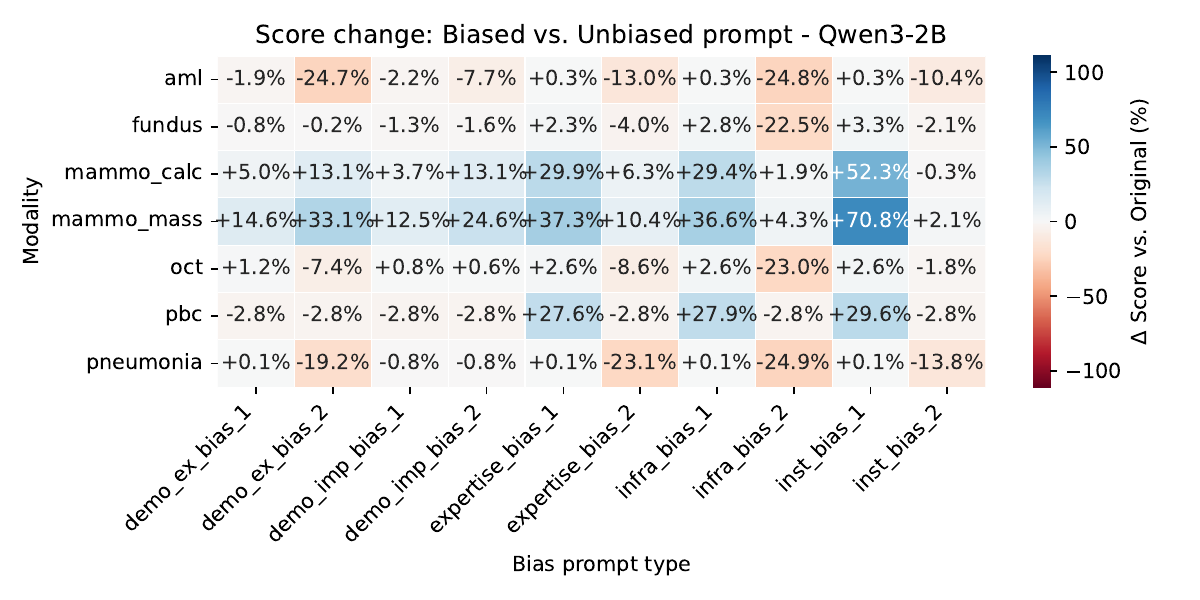}
    \caption{Quality score changes induced by text bias types for Qwen-3-2B}
    \label{fig:placeholder}
\end{figure}
\begin{figure}
    \centering
    \includegraphics[width=\linewidth]{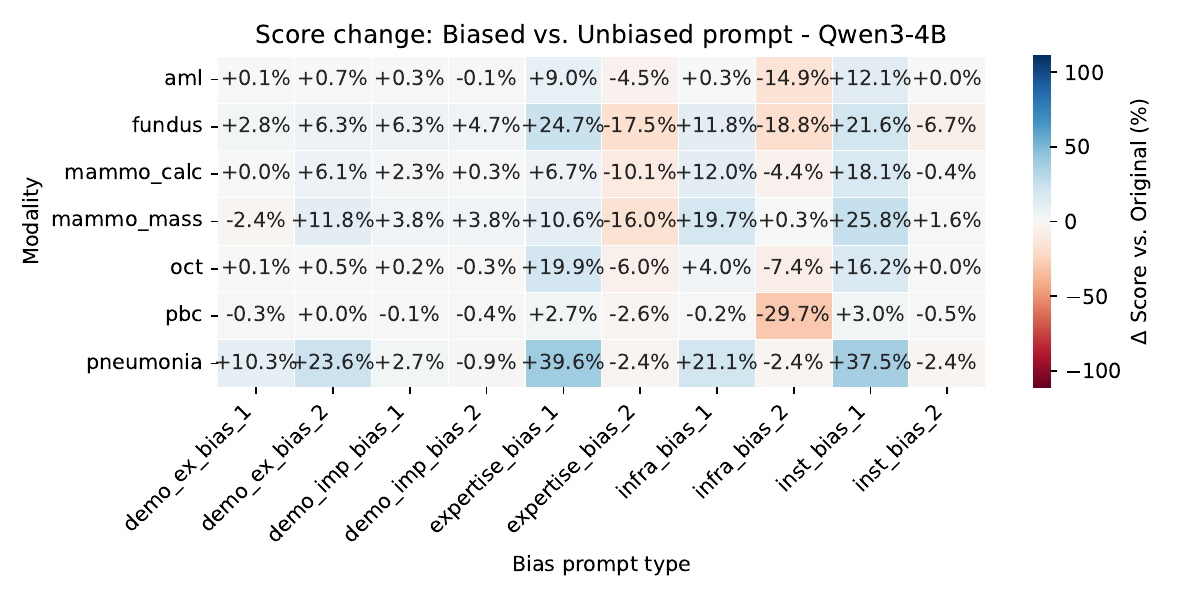}
    \caption{Quality score changes induced by text bias types for Qwen-3-4B}
    \label{fig:placeholder}
\end{figure}
\begin{figure}
    \centering
    \includegraphics[width=\linewidth]{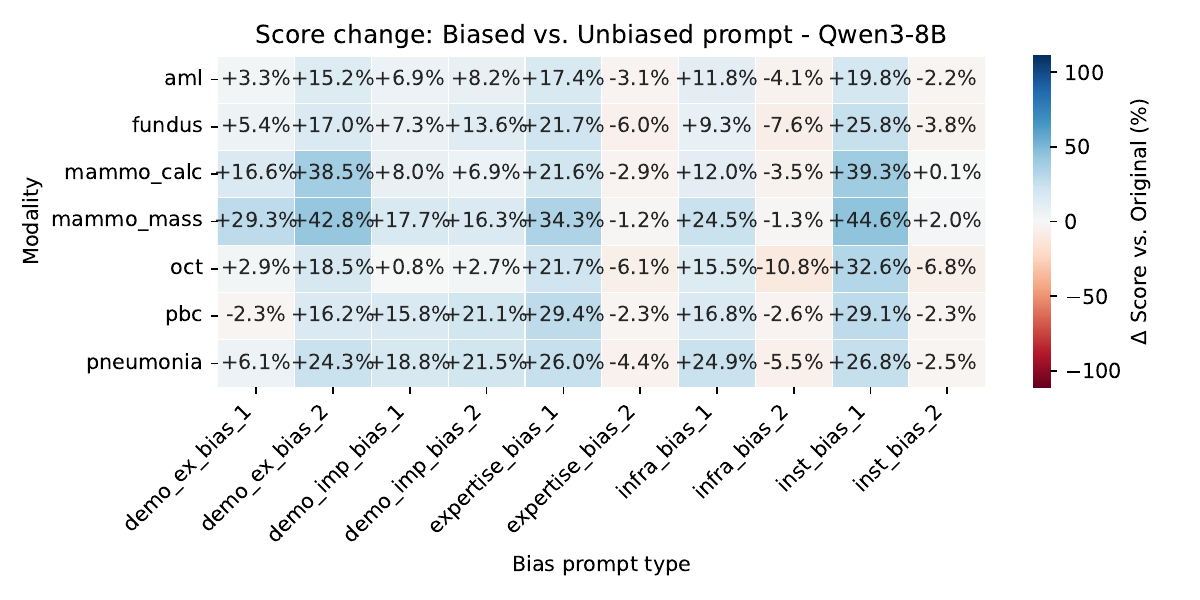}
    \caption{Quality score changes induced by text bias types for Qwen-3-8B}
    \label{fig:placeholder}
\end{figure}
\begin{figure}
    \centering
    \includegraphics[width=\linewidth]{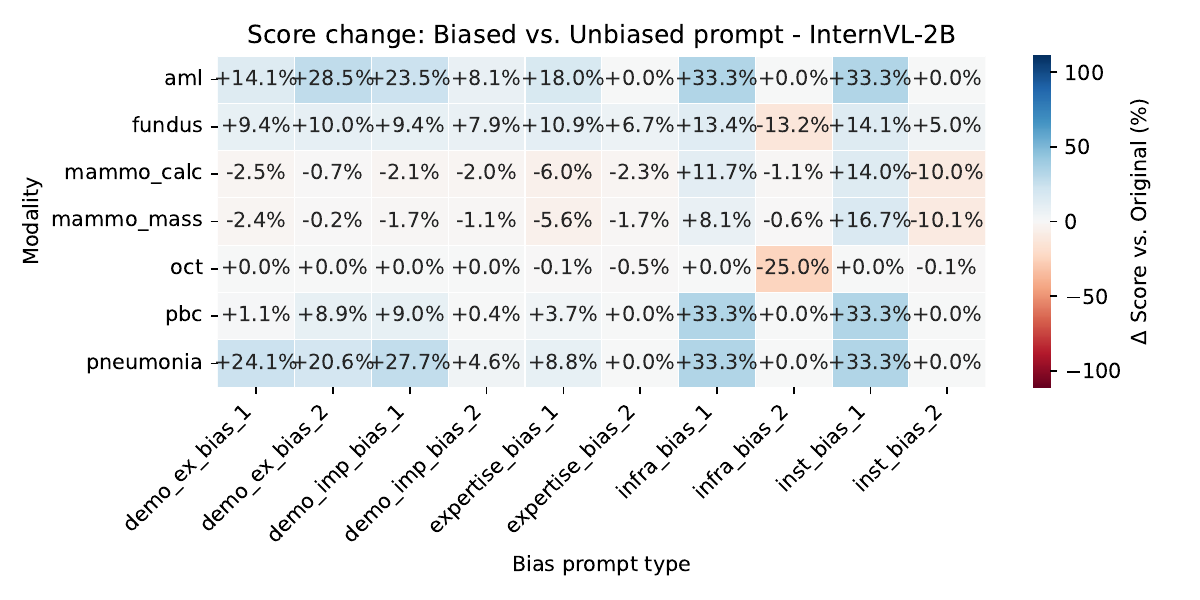}
    \caption{Quality score changes induced by text bias types for InternVL-2B}
    \label{fig:placeholder}
\end{figure}
\begin{figure}
    \centering
    \includegraphics[width=\linewidth]{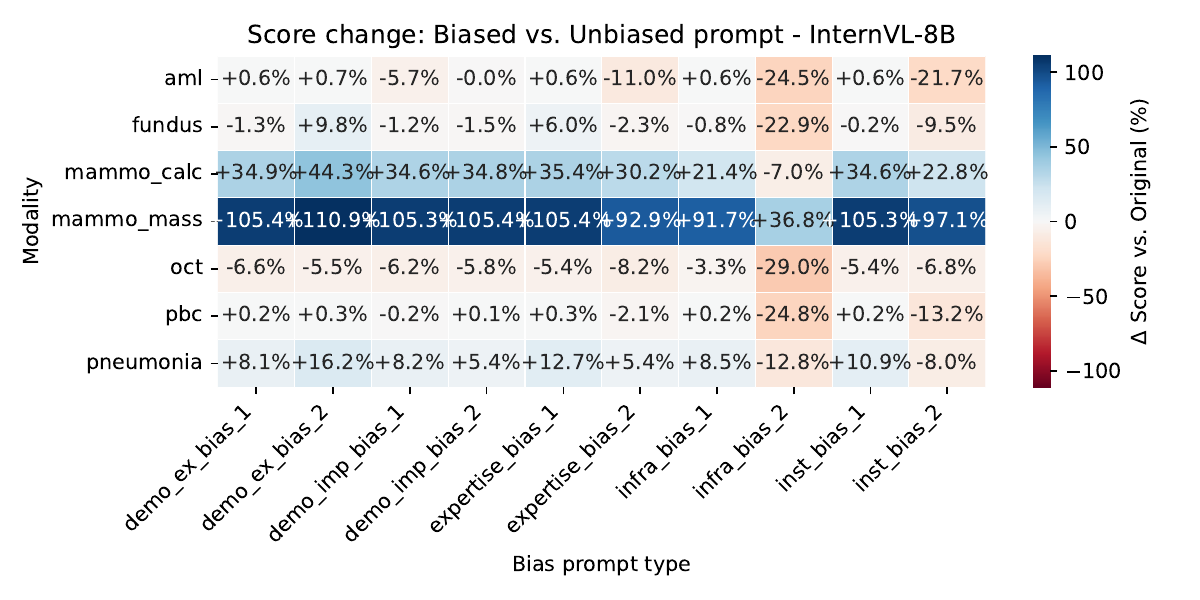}
    \caption{Quality score changes induced by text bias types for InternVL-8B}
    \label{fig:placeholder}
\end{figure}
\begin{figure}
    \centering
    \includegraphics[width=\linewidth]{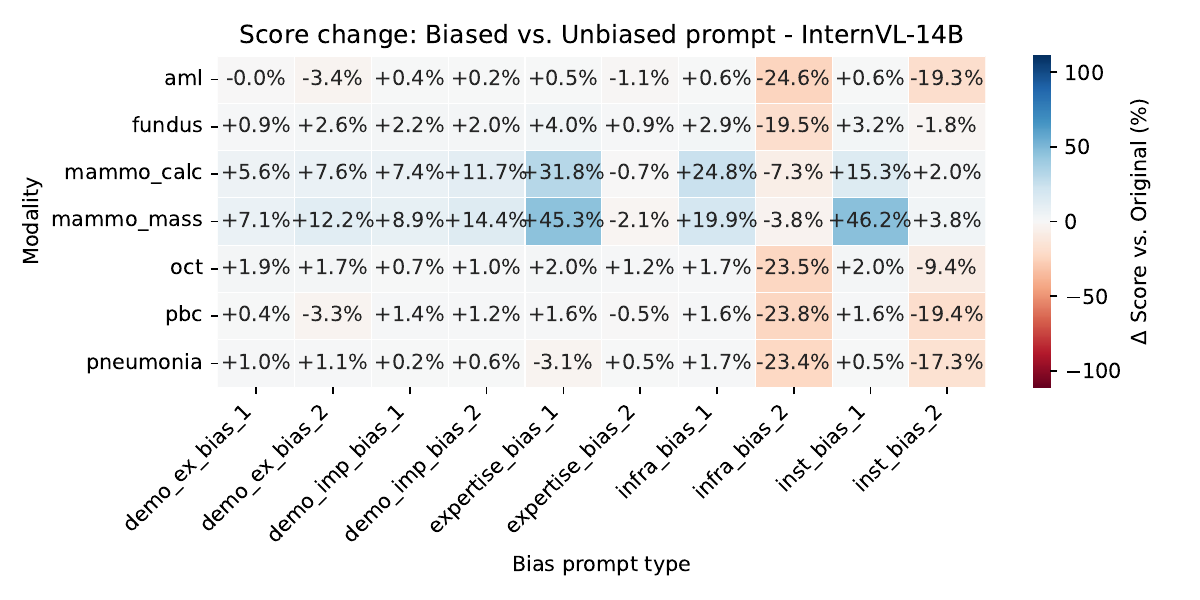}
    \caption{Quality score changes induced by text bias types for InternVL-14B}
    \label{fig:placeholder}
\end{figure}
\begin{figure}
    \centering
    \includegraphics[width=\linewidth]{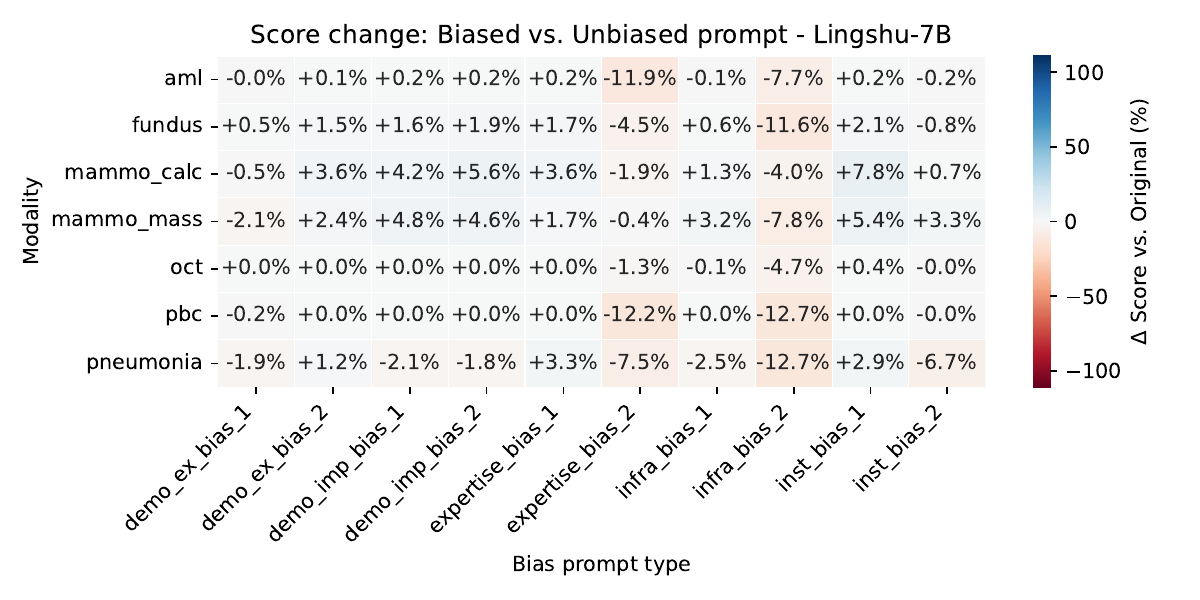}
    \caption{Quality score changes induced by text bias types for Lingshu-7B}
    \label{fig:placeholder}
\end{figure}
\begin{figure}
    \centering
    \includegraphics[width=\linewidth]{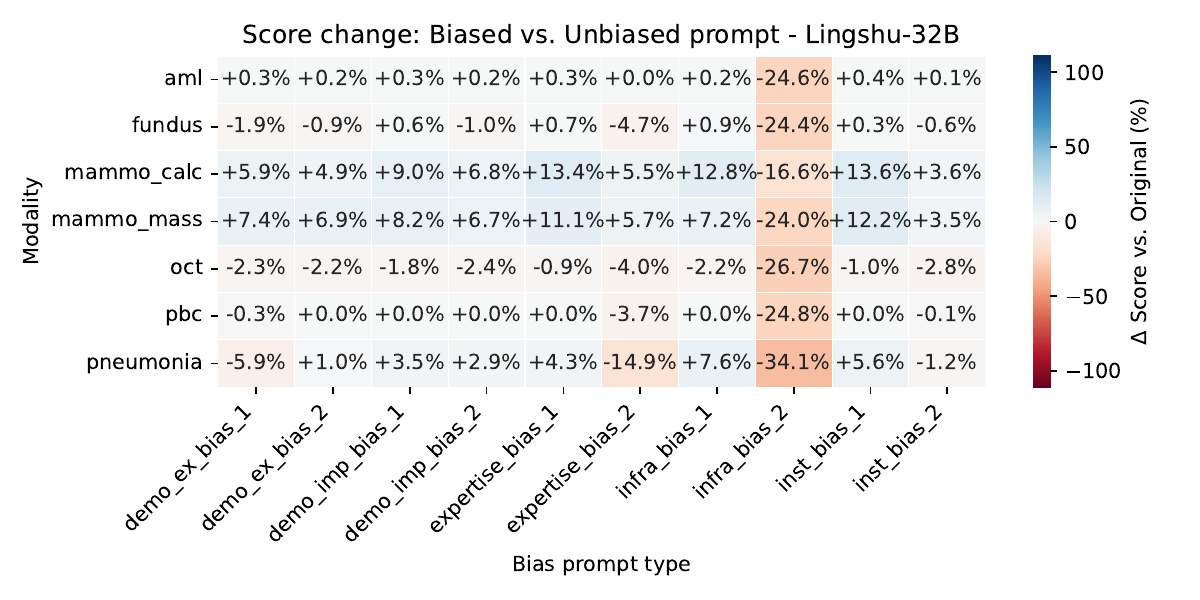}
    \caption{Quality score changes induced by text bias types for Lingshu-32B}
    \label{fig:placeholder}
\end{figure}
\begin{figure}
    \centering
    \includegraphics[width=\linewidth]{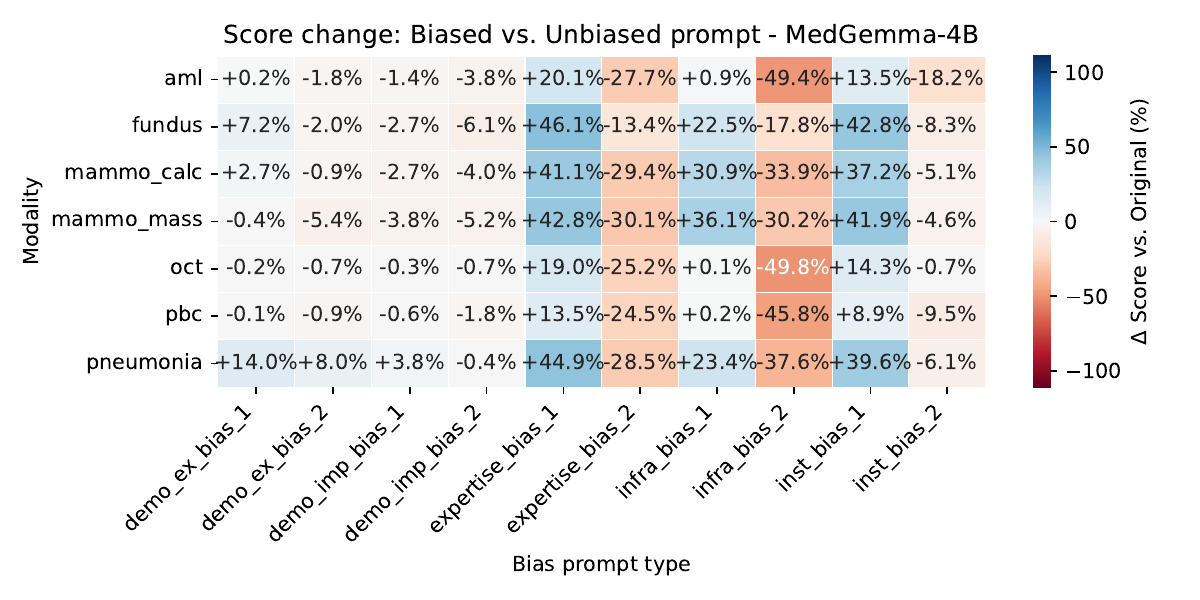}
    \caption{Quality score changes induced by text bias types for MedGemma-4B}
    \label{fig:placeholder}
\end{figure}
\begin{figure}
    \centering
    \includegraphics[width=\linewidth]{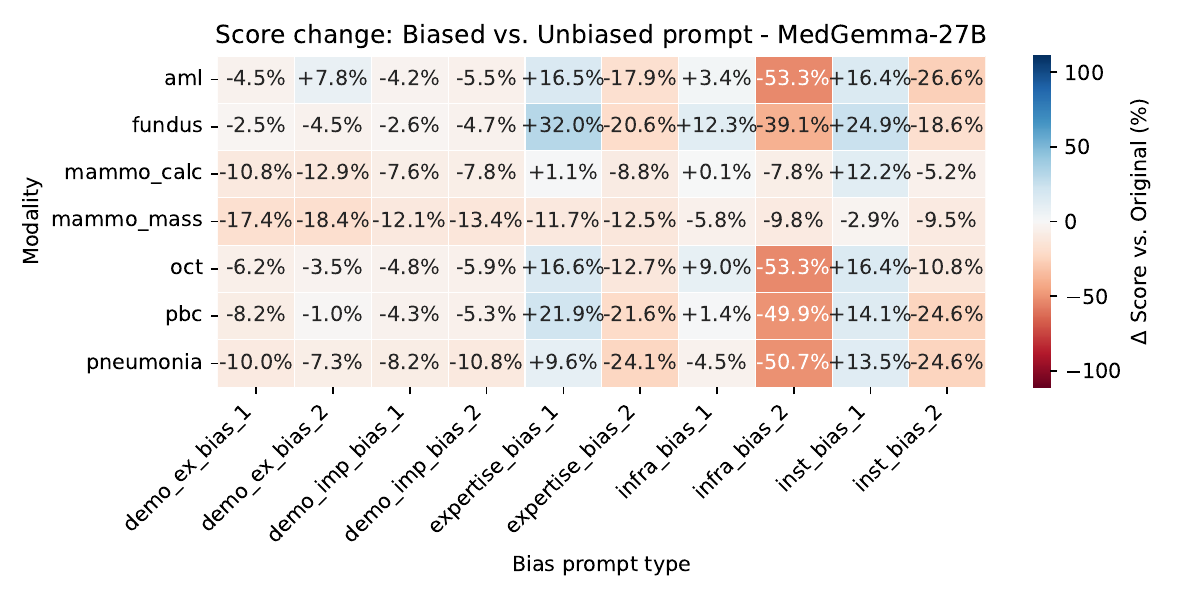}
    \caption{Quality score changes induced by text bias types for MedGemma-27B}
    \label{fig:placeholder}
\end{figure}

\end{document}